\documentclass{article}

\PassOptionsToPackage{dvipsnames}{xcolor}
\usepackage[allcolors=NavyBlue,colorlinks=true]{hyperref}       %

\usepackage[accepted]{icml2024}

\PassOptionsToPackage{numbers, compress}{natbib}

\usepackage{multirow}
\usepackage{threeparttable}
\usepackage{diagbox}
\usepackage[utf8]{inputenc} %
\usepackage[T1]{fontenc}    %
\usepackage{url}            %
\usepackage{booktabs}       %
\usepackage{amsfonts}       %
\usepackage{nicefrac}       %
\usepackage{microtype}      %
\usepackage{xspace}
\usepackage{caption,subcaption}

\usepackage{wrapfig}

\usepackage{algorithm}
\usepackage[noend]{algpseudocode}

\usepackage{comment}
\usepackage{prettyref}
\usepackage{amsfonts}
\usepackage{makecell}
\usepackage{adjustbox}
\usepackage{multicol}

\usepackage{amsmath}
\usepackage{amssymb}
\usepackage{mathtools}
\usepackage{amsthm}

\usepackage[
textsize=scriptsize,
]{todonotes}

\usepackage{xspace}

\usepackage{titletoc}
\usepackage[toc,page,header]{appendix}
\usepackage{minitoc}

\definecolor{colorcomment}{RGB}{160, 190, 210}%
\makeatletter
\algnewcommand{\LineComment}[1]{\Statex \hskip\ALG@thistlm \(\triangleright\) 
{\color{colorcomment}#1}}
\makeatother

\makeatletter
\algnewcommand{\IndentLineComment}[1]{\Statex \hskip\ALG@tlm \(\triangleright\) {\color{colorcomment}#1}}
\makeatother

\icmltitlerunning{Entropy-Reinforced Planning with Large Language Models for Drug Discovery}

\begin{document}

\twocolumn[

\icmltitle{Entropy-Reinforced Planning with Large Language Models for Drug Discovery}

\icmlsetsymbol{equal}{*}

\begin{icmlauthorlist}
\icmlauthor{Xuefeng Liu}{uchicago,equal}
\icmlauthor{Chih-chan Tien}{uchicago,equal}
\icmlauthor{Peng Ding}{uchicago}
\icmlauthor{Songhao Jiang}{uchicago}
\icmlauthor{Rick L. Stevens}{uchicago,argonne}
\end{icmlauthorlist}

\icmlaffiliation{uchicago}{Department of Computer Science, University of Chicago, Chicago, IL, USA}
\icmlaffiliation{argonne}{Argonne National Laboratory, Lemont, IL, USA}

\icmlcorrespondingauthor{Xuefeng Liu}{xuefeng@uchicago.edu}

\icmlkeywords{Machine Learning, ICML}

\vskip 0.3in
]

\printAffiliationsAndNotice{\icmlEqualContribution} %

\newcommand*\boldell{\pmb{\ell}}

\newcommand\lossIL{e}
\newcommand\loss{\ell}
\newcommand\lossB{\ensuremath{\boldell}}

\newcommand\Loss{\ensuremath{L}}
\newcommand\LossB{\ensuremath{\mathbi{L}}}

\newcommand\Policies{\ensuremath{\Pi}}
\newcommand\policy{\ensuremath{\pi}}
\newcommand\PoliciesDist{q}
\newcommand\PoliciesDistB{\mathbi{q}}
\newcommand\state{s}
\newcommand\stateB{\ensuremath{\hat{{\mathbi{\state}}}}}
\newcommand\stateDist{d}
\newcommand\stateDistB{\mathbi{d}}
\newcommand\transDynamics{\mathcal{P}}
\newcommand\reward{r}
\newcommand\RFunc{R}

\newcommand\DFunc{D}
\newcommand\Gradient{{g}}

\newcommand\VFunc{\ensuremath{\ensuremath{V}}}
\newcommand\Func{\ensuremath{\ensuremath{f}}}
\newcommand\QFunc{\ensuremath{\ensuremath{Q}}}
\newcommand\Adv{\mathbi{A}}
\newcommand\AFunc{\mathbi{A}}
\newcommand\pw{h}
\newcommand\pwB{\mathbi{h}}
\newcommand\PolicyAdviceMatrix{\mathcal{E}^{(t)}}

\newcommand\Indicator{\textit{I}}
\newcommand\queryProb{z}

\newcommand{\trollin}{\psi}

\def\mathbi#1{\textbf{\em #1}}
\newcommand\ipm{\mathbi{I}}
\newcommand\Evaluators{\Policies}
\newcommand\evaluator{\policy}
\newcommand\policyIndex{k}
\newcommand\modelIndex{j}
\newcommand\ModelsDist{w}%
\newcommand\ModelsDistB{\mathbi{w}}%

\newcommand\pd{\ensuremath{\hat{\clabel}}}
\newcommand\pdB{\ensuremath{\hat{{\mathbi{\clabel}}}}}

\newcommand\discFactor{\gamma}

\newcommand\qlb{\ensuremath{{\frac{1}{\sqrt{t}}}}}
\newcommand\constant{\ensuremath{C}}

\newcommand\Models{\ensuremath{\mathcal{F}}}
\newcommand\TasksNum{\ensuremath{k}}

\newcommand\KAhead{\ensuremath{\text{\emph{e}-steps ahead confidence}}}

\newcommand\ONum{\ensuremath{K}}

\newcommand\Num{\ensuremath{N}}

\newcommand\model{\ensuremath{f}}
\newcommand\instance{\ensuremath{\mathbi{x}}}
\newcommand\domInstance{\ensuremath{\mathcal{X}}}
\newcommand\domClabel{\ensuremath{\mathcal{Y}}}
\newcommand\stateSpace{\ensuremath{\mathcal{S}}}
\newcommand\action{\ensuremath{a}}
\newcommand\actionDist{\ensuremath{\mathbi{A}}}
\newcommand\actionSpace{\ensuremath{\mathcal{A}}}
\newcommand\trajectory{\ensuremath{\mathcal{\tau}}}
\newcommand\horizon{\ensuremath{H}}
\newcommand\ENum{\ensuremath{N}} %
\newcommand\RNum{\ensuremath{N}} %
\newcommand\eNum{\ensuremath{n}} %

\newcommand\UCB{\ensuremath{UCB}} 
\newcommand\PUCB{\ensuremath{P-UCB}} 
\newcommand\PHUCB{\ensuremath{\text{P}\mathcal{H}\text{-UCB}}}
\newcommand\PHUCT{\ensuremath{\text{P}\mathcal{H}\text{-UCT}}}

\newcommand\Reward{\ensuremath{\mathcal{R}}} %
\newcommand\visit{\ensuremath{{v}}}
\newcommand\PVisit{\ensuremath{\mathcal{V}}} %
\newcommand\wNum{\ensuremath{{t}}} %

\newcommand\weight{\ensuremath{{w}}} %

\newcommand\rationale{\ensuremath{\mathcal{S}}}

\newcommand\molecular{\ensuremath{\mathcal{G}}}

\newcommand\threshold{\ensuremath{{\epsilon}}}

\newcommand\exploreProb{\ensuremath{{p}}}
\newcommand\dataset{\ensuremath{\mathcal{D}}}

\newcommand\KPUCT{\ensuremath{{\text{P-UCT-K}}}}
\newcommand\BS{\ensuremath{{\text{Beam-Search}}}}

\newcommand\RewardFunc{\ensuremath{r}}

\newcommand\Regret{\ensuremath{R}}
\newcommand\TotalExpCost{\ensuremath{J}}
\newcommand\clabel{y} %
\newcommand\clabelB{\mathbi{y}} %
\newcommand\Learner{\ensuremath{\mathcal{AL}}\xspace} 

\newcommand\MDP{\ensuremath{\mathcal{M}}}

\newcommand\adviceMatrix{\ensuremath{\mathbi{E}}}
\newcommand\adviceMatrixI{\ensuremath{\mathcal{E}}}%
\newcommand\adviceMatrixIB{\ensuremath{\mathbb{\mathcal{E}}}}
\newcommand\expectation{\ensuremath{\mathbb{E}}}
\newcommand\budget{\ensuremath{b}}

\newcommand\entropyF{\mathfrak{E}}

\newcommand\minpgap[1]{\ensuremath{\rho_{#1}}}

\newcommand\maxind[1]{\ensuremath{\text{maxind}(#1)}}

\renewcommand{\Pr}[1]{\ensuremath{\mathbb{P}\left[#1\right] }}
\newcommand{\Prover}[2]{\ensuremath{\mathbb{P}_{#1}\!\left[#2\right]}}

\newcommand{\expct}[1]{\mathbb{E}\left[#1\right]}
\newcommand{\expctover}[2]{\mathbb{E}_{#1}\!\left[#2\right]}
\newcommand\RationaleBuffer{\ensuremath{\mathcal{B^R}}}
\newcommand\MC{\ensuremath{\mathcal{MC}}}

\newcommand\critic{\ensuremath{{C}}\xspace} 
\newcommand\Buffer{\ensuremath{\mathcal{B}}}

\newcommand{\abs}[1]{\left\vert#1\right\vert}
\def \argmax {\mathop{\rm arg\,max}}
\def \argmin {\mathop{\rm arg\,min}}

\newcommand\Gen{\ensuremath{\mathcal{G}_{\theta}}}
\newcommand\Dis{\ensuremath{\mathcal{D}_{\phi}}}

\newcommand\fm{\ensuremath{\mathbb{\psi}}}

\newcommand{\algname}{\textsc{ERP}\xspace}
\newcommand{\LLM}{\textsc{LLMs}\xspace}
\newcommand{\TOPK}{\textsc{Top-K}\xspace}
\newcommand{\TOPP}{\textsc{Top-P}\xspace}
\newcommand{\TOPPK}{\textsc{Top-PK}\xspace}
\newcommand{\hypothesis}{\mathcal{Y}}

\newcommand{\algRecommend}{\textsc{Recommend}\xspace}

\newif\iffinal
\finaltrue

\iffinal
    \newcommand{\reb}[1]{{\color{red} #1}} %
    \newcommand{\fix}[1]{#1}
    \newcommand{\XL}[1]{}
    \newcommand{\XLinline}[1]{}
    \newcommand{\SH}[1]{}
    \newcommand{\SHinline}[1]{}
    \newcommand{\CT}[1]{}
    \newcommand{\CTinline}[1]{}
    \newcommand{\DP}[1]{}
    \newcommand{\DPinline}[1]{}
    \newcommand{\note}[1]{}
    \newcommand{\pref}[1]{}
\else
    \newcommand{\reb}[1]{{\color{red} #1}} %
    \newcommand{\fix}[1]{{\color{red} #1}}
    \newcommand{\SH}[1]{\todo[fancyline,color=Maroon!40]{SH: #1}\xspace}
    \newcommand{\DP}[1]{\todo[fancyline,color=Purple!40]{DP: #1}\xspace}
    \newcommand{\XL}[1]{\todo[fancyline,color=NavyBlue!40]{XL: #1}\xspace}
    \newcommand{\XLinline}[1]{\textcolor{ForestGreen}{[XL: #1]}}
    \newcommand{\CT}[1]{\todo[fancyline,color=NavyBlue!40]{CT: #1}\xspace}
    \newcommand{\CTinline}[1]{\textcolor{NavyBlue}{[CT: #1]}}
    \newcommand{\FF}[1]{\todo[fancyline,color=ForestGreen!40]{FX: #1}\xspace}
    \newcommand{\note}[1]{{\color{purple}[XL: #1]}}
    \newcommand{\question}[1]{\todo[fancyline,color=Maroon!40]{XL: #1}\xspace}
    \newcommand{\pref}[1]{{\color{blue}(\ref{#1})}}
\fi

\newcommand{\NonNegativeReals}{\ensuremath{\mathbb{R}_{\ge 0}}}
\newcommand{\PositiveIntegers}{\ensuremath{\mathbb{Z^+}}}
\newcommand{\integers}{\ensuremath{\mathbb{Z}}}
\newcommand{\nats}{\ensuremath{\mathbb{N}}}
\newcommand{\reals}{\ensuremath{\mathbb{R}}}
\newcommand{\rationals}{\ensuremath{\mathbb{Q}}}
\newcommand{\distrib}[0]{\ensuremath{\mathcal{D}}}
\newcommand{\matroid}{\ensuremath{\mathcal{M}}}

\newcommand{\defref}[1]{Definition~\ref{#1}}
\newcommand{\tabref}[1]{Table~\ref{#1}}
\newcommand{\figref}[1]{Fig~\ref{#1}}
\newcommand{\eqnref}[1]{\text{Eq.}~(\ref{#1})}
\newcommand{\secref}[1]{\S\ref{#1}}
\newcommand{\appref}[1]{Appendix~\ref{#1}}
\newcommand{\thmref}[1]{Theorem~\ref{#1}}
\newcommand{\corref}[1]{Corollary~\ref{#1}}
\newcommand{\propref}[1]{Proposition~\ref{#1}}
\newcommand{\lemref}[1]{Lemma~\ref{#1}}
\newcommand{\conref}[1]{Condition~\ref{#1}}
\newcommand{\assref}[1]{Assumption~\ref{#1}}
\newcommand{\egref}[1]{Example~\ref{#1}}
\newcommand{\algoref}[1]{Algorithm~\ref{#1}}

\newcommand{\op}[1]{\operatorname{#1}}
\newcommand{\paren} [1] {\ensuremath{ \left( {#1} \right) }}
\newcommand{\parenb} [1] {\ensuremath{ \big( {#1} \big) }}
\newcommand{\bigparen} [1] {\ensuremath{ \Big( {#1} \Big) }}
\newcommand{\biggparen} [1] {\ensuremath{ \bigg( {#1} \bigg) }}
\newcommand{\Biggparen} [1] {\ensuremath{ \Bigg( {#1} \Bigg) }}
\newcommand{\bracket}[1]{\left[#1\right]}
\newcommand{\tuple}[1]{\ensuremath{\left\langle #1 \right\rangle}}
\newcommand{\curlybracket}[1]{\ensuremath{\left\{#1\right\}}}
\newcommand{\condcurlybracket}[2]{\ensuremath{\left\{#1\left\lvert\:#2\right.\right\}}}

\theoremstyle{plain}
\newtheorem{theorem}{Theorem}[section]
\newtheorem{proposition}[theorem]{Proposition}
\newtheorem{lemma}[theorem]{Lemma}
\newtheorem{corollary}[theorem]{Corollary}
\theoremstyle{definition}
\newtheorem{definition}[theorem]{Definition}
\newtheorem{assumption}[theorem]{Assumption}
\theoremstyle{remark}
\newtheorem{remark}[theorem]{Remark}

\newcommand{\stochastic}{\textcolor{ForestGreen}{\textsc{stochastic}}}
\newcommand{\adversarial}{\textcolor{Maroon}{\textsc{adversarial}}}
\newcommand{\Cost}{C}

\newcommand{\numTotalPolicies}{m}

\newcommand{\bigO}{O}

\begin{abstract}

The objective of drug discovery is to identify chemical compounds that possess specific pharmaceutical properties toward a binding target. Existing large language models (\LLM) can achieve high token matching scores in terms of likelihood for molecule generation.
However, relying solely on LLM decoding often results in the generation of molecules that are either invalid due to a single misused token, {or} suboptimal due to unbalanced exploration and exploitation as a consequence of the LLM's prior experience.
Here we propose \algname, Entropy-Reinforced Planning for Transformer Decoding, which employs an entropy-reinforced planning algorithm to enhance the {Transformer} decoding process and strike a balance between exploitation and exploration.
\algname aims to achieve improvements in multiple properties compared to direct sampling from the Transformer.
We evaluated \algname on the SARS-CoV-2 virus (3CLPro) and human cancer cell target protein (RTCB) benchmarks and {demonstrated that, in both benchmarks, \algname consistently outperforms the current state-of-the-art algorithm by 1-5 percent, and baselines by 5-10 percent, respectively.} {Moreover, such improvement is robust across Transformer models trained with different objectives. } 
\fix{Finally, to further illustrate the capabilities of \algname, we tested our algorithm on three code generation benchmarks and outperformed the current state-of-the-art approach as well.} \fix{Our code is publicly available at: \url{https://github.com/xuefeng-cs/ERP}.}

\end{abstract}

\section{Introduction}

The emergence and rapid evolution of COVID-19~\citep{yuki2020covid,hadj2022covid} has created an urgent need for the expedited discovery of effective drugs---a process in which approaches based on large language models (\LLM) have begun to play a pivotal role \citep{frey2023neural,bagal2021molgpt}.

\fix{Transformer~\citep{vaswani2017attention} and Transformer-based} LLMs such as GPT~\citep{brown2020language,ouyang2022training,achiam2023gpt} and BERT~\citep{devlin2018bert} have shown remarkable success in diverse fields, from robotics~\citep{vemprala2023chatgpt} to education~\citep{tang2023science} and speech recognition~\citep{yang2021multi}. However, LLM applications in drug discovery and molecular design have not yet achieved comparable levels of performance---shortcomings that we attribute to the challenges of generating molecular sequences that both are valid (difficult given that even a single erroneous token can lead to complete failure) and meet multiple other criteria simultaneously.

A consequence of these challenges is that existing LLM-based molecular discovery algorithms
\citep{frey2023neural,bagal2021molgpt} 
must sample inordinate numbers of molecular structures. 
The primary factor contributing to this sample inefficiency is the use of the Transformer beam search algorithm.
While often an effective heuristic for text generation \citep{meister2020if}, beam search has two major deficiencies when used to generate molecular structures: it can neither terminate generation early if it detects that the molecular sequence being produced is likely to fail, nor guide the generation process toward structures with superior pharmaceutical properties. 

Meanwhile, planning algorithms like Monte Carlo Tree Search (MCTS) have shown promise for \fix{decision making problems}. Since the introduction of the Upper Confidence Bound applied to Trees (UCT) method \citep{kocsis2006bandit}, MCTS has demonstrated exceptional performance in various domains, notably AlphaGo \citep{silver2017mastering}, where MCTS was integrated with deep neural networks (DNNs). However, when applied to molecular design it still incurs a non-trivial search cost on unpromising molecules and thus also suffers from %
sample inefficiency.

Given the vast search space for molecular sequences, relying solely on the MCTS planner or combining it with conventional DNN approaches may not identify high-reward molecules efficiently. This is where \fix{we demonstrate} the pre-trained molecular Transformer proves to be invaluable \fix{for molecular design}\XL{define what a pre-trained molecular transformer is.}. Recently, Planning-Guided Transformer Decoding (PG-TD) \citep{zhang2023planning} has employed the MCTS planning algorithm to guide the Transformer decoder for code generation with a single pass/fail reward function. However, PG-TD adapts the Transformer decoder's next token selection, which might lead to the over-exploitation of certain tokens while failing to explore regions of uncertainty where the true optimal solution might be hidden.
In this study, we ask: How can we improve the exploration and exploitation mechanisms in the MCTS planning-guided Transformer decoding process for multiple properties enhancement in drug discovery?

\fix{To address this challenge,} we introduce a novel algorithm, \textbf{E}ntropy-\textbf{R}einforced \textbf{P}lanning for Transformer Decoding, or \algname, which integrates an entropy \fix{based}
planning 
algorithm into the Transformer-based generation process. Specifically, for the selection procedure, \algname goes beyond pure MCTS, which only considers the visitation numbers of parent and child nodes in the bonus term. \fix{Instead, ERP} applies the Transformer decoder for the next token probability and an \emph{e}-step forward entropy measurement term to evaluate the potential uncertainty following the token. Moreover, the entropy-reinforced MCTS planner directs the Transformer decoder towards exploring promising underlying molecular spaces, instead of merely optimizing the likelihood of the sequences it generates, and also enables the creation of more controllable candidate molecules.
Specifically, ERP incorporates the Transformer's next token probabilities and the $\KAhead$ method for assessing token uncertainties into the MCTS planner's Upper Confidence Bound (UCB)-based exploration bonus function. In so doing, it increases the efficiency of sampling sequences that yield higher rewards. {Experiments show that the incorporation next token probability and the \emph{e}-step forward entropy measurement boosts the normalized reward by 5-10 percent. } In the expansion stage, \algname uses both the \TOPK and $\TOPP$ approaches of the Transformer to select the most likely next token given the current state, improving sample efficiency by excluding non-promising action spaces. For the evaluation stage, we employ the Transformer's beam search to estimate the reward for an incomplete, partial molecular sequence and then backpropagate the value to ancestor nodes.

\vspace{1ex}
\noindent
\textbf{Our contributions} are as follows:

We present a novel algorithm, \algname, which employs an Entropy-Reinforced Planner to guide the decoding process of a pretrained Transformer. This algorithm combines LLMs and planning algorithms to accelerate drug discovery by enhancing the quality, diversity, and sample efficiency of generated molecules.

We introduce a novel selection algorithm, \PHUCT,
that takes into account the Transformer's next token probability and incorporates an \emph{e}-step forward entropy measurement for token selection. These methods reduce uncertainty and improve exploration and exploitation in decision-making.

Lastly, we conduct extensive experiments on the \fix{3CLPro (PDBID: 7BQY) SARS-CoV-2 protein and RTCB (PDBID: 4DWQ) human cancer cell protein} benchmarks to compare \algname with PG-TD \citep{zhang2023planning}, UCT~\citep{kocsis2006bandit}, beam-search, and sampling algorithm from baseline Transformer models. 
Our results show that \algname outperforms the current state-of-the-art (PG-TD) as well as all baseline models across these experiments. Moreover,  \algname can leverage general pretrained models to improve sample efficiency, correct biased models through controlled generation, and achieve continuous improvement over reinforcement learning (RL) well-\fix{fine-tuned} 
Transformer models by balancing exploitation and exploration for potentially higher-reward molecular spaces.
\fix{To further illustrate the capabilities of \algname, we conducted experiments on three code generation benchmarks and demonstrated its superiority compared to the current state-of-the-art approach.}

\section{Related Work}\label{sec:related}

We review related work in LLM-based drug discovery, planning and RL for molecule generation, interactive imitation learning, and planning in natural language generation.

\subsection{LLM-based drug discovery}

LLM generative capabilities have been applied to molecule generation \citep{bagal2021molgpt,rothchild2021c5t5,wang2022transformer} in works like MolGPT \citep{bagal2021molgpt}, ChemGPT \citep{frey2023neural} and C5T5 \citep{rothchild2021c5t5}, and to drug discovery \citep{liu2023chatgpt,seidl2023enhancing}, with molecules represented via a linear string representation, such as simplified molecular-input line-entry system (SMILES) \citep{weininger1988smiles} or 
self-referencing embedded strings (SELFIES) \citep{krenn2020self}. 
These approaches have at times demonstrated performance on par with conventional methods that use traditional molecular representations, with the potential to provide distinctive predictive outcomes. Nevertheless, they remain limited to basic molecular transformations and are only capable of supporting roles in molecular design \fix{procedures}.  As noted by \citet{murakumo2023llm}, the direct creation of molecules through pre-trained LLMs generally results in less impressive outcomes. 
In contrast, our approach integrates a tree search planning algorithm and multiple critics to steer pre-trained LLMs toward generating higher-quality molecules.

\subsection{Planning and RL for molecule generation}

Several researchers \citep{yang2017chemts, yang2020practical, yoshizawa2022selective, m2017mdts, hong2023retrosynthetic} have integrated planning algorithms like MCTS into molecule generation.
However, those workers focused on objectives outside of drug discovery, such as material science; eschewed (in order to enhance sample efficiency) the use of LLMs; and/or overlooked incorporating confidence in their exploratory processes.
Online RL has been applied to improve molecule generation, both on SMILES string representations \citep{born2021paccmannrl, guimaraes2017objective, neil2018exploring, olivecrona2017molecular, popova2018deep,  staahl2019deep, tan2022drlinker, wang2022reinforcement, zhang2023universal, zhou2019optimization} and graph-based representations
\citep{atance2022novo,gottipati2020learning, jin2020multi,wu2022rlcg,you2018graph}.
As we describe in the following, we employ RL but with
guided exploration by an MCTS planner.
In this regard, our problem setting aligns more closely with imitation learning, which utilizes an expert to guide the learning process.

\subsection{Interactive imitation learning}

Imitation learning (IL) is a machine learning technique in which an agent learns to perform tasks by observing and mimicking the actions of an expert. Unlike pure RL, which struggles with large action and state spaces, IL has shown to be more sample-efficient and effective in environments with sparse rewards due to its ability to leverage expert guidance \citep{ross2011reduction}. Interactive IL methods such as DAgger \citep{ross2014reinforcement}, MAPS \citep{liu2023active}, and RPI \citep{liu2023blending} employ a Roll-in-Roll-out (RIRO) strategy to provide guidance. The learner first performs a default roll-in from the initial state and then adopts the expert's guidance to roll out the subsequent steps in the trajectory, thereby correcting the learner's behavior. In contrast, we use an MCTS planner to guide the whole decoding process of the Transformer and rely on the Transformer's beam search to complete the trajectory by forming a whole molecule. Instead of engaging the \fix{learner only in the roll-in phase and} expert only during the roll-out phase, we involve the Transformer decoder in both the roll-in and roll-out phases.

\subsection{Planning in natural language generation}

\citet{scialom2021beam}, \citet{leblond2021machine}, and \citet{chaffin2021ppl}, among others, have applied the MCTS planning algorithm to optimize text outputs for various NLP tasks. PG-TD \citep{zhang2023planning} focuses on code generation using a single reward function (pass or fail). Our work here is the first to integrate a tree search planning algorithm with an LLM decoder specifically for de novo drug discovery. Unlike previous approaches that employed pretrained LLMs without evaluating the potential certainty of each generated token and that optimized for only a single objective, we introduce an \emph{e}-step forward entropy measurement, with the reward computed based on multiple criteria.
\fix{We show that by combining the entropy-reinforced planner with the Transformer decoder, our algorithm balances exploration and exploitation in the molecular space and thus discovers molecules with high rewards.}

\section{Preliminaries}
We describe LLM, MCTS, and drug discovery along with their mathematical notation in the subsequent sections and unify them under the Markov decision processes framework.

\textbf{Markov decision processes.} We consider a finite-horizon Markov decision process (MDP)~\citep{puterman2014markov} $\MDP_0 = \langle \stateSpace, \actionSpace,\horizon,\transDynamics,\RFunc \rangle$ where $\stateSpace$ is a finite set of states; 
$\actionSpace$ is a finite set of actions;
$\horizon$ represents the planning horizon;
$\transDynamics$ is the deterministic transition dynamics $\transDynamics: \stateSpace \times \actionSpace \rightarrow \stateSpace'$, which concatenates a state $\state$ with a token $\action$, and an episode ends when the agent performs the termination action; and $\RFunc: \stateSpace \times \actionSpace \rightarrow \mathbb{R}$ is a reward function, which only scores a complete molecule (the reward of a partial molecule is 0).
The initial state distribution is \fix{$\stateDist_0$}. 
The policy $\pi: \mathcal{S} \rightarrow \Delta(\mathcal{A})$ assigns a distribution over actions based on the current state. The $Q$-value function, expressed as $Q^{\pi}:\mathcal{S} \times \mathcal{A} \rightarrow \mathbb{R}$, evaluates the performance of the policy.

\textbf{LLM.} We represent each molecule as a state $\state$ comprising a start token 
$\bracket{\text{BOS}}$, a SMILES~\citep{weininger1988smiles} string, and (for both partial and complete molecules) a termination action $\bracket{\text{EOS}}$. 
All possible molecules form the state space $\stateSpace$.
\fix{Let $\hypothesis_t$ be the hypothesis space in step $t$ (sequence length $t$). We have $\hypothesis_t \subseteq \stateSpace_t \subseteq \stateSpace|_{t\in \bracket{\horizon}}$.}
We define the set of complete hypotheses, i.e., complete molecules, as
\begin{equation}
    \hypothesis_{\horizon} := \curlybracket{\text{[BOS]} \circ \mathbf{v} \circ \text{[EOS]}~|~\mathbf{v}\in \mathcal{V}^*},
\end{equation}
where $\mathcal{V}^*$ represents the Kleene closure of $\mathcal{V}$ and $\circ$ denotes string concatenation.
Each action $\action \in \actionSpace$ is represented as token $y$ in the Transformer's vocabulary $\mathcal{V}$,  $\mathcal{V}:=\actionSpace$.
Given a set of molecules $\Buffer$, we train the LLM generator policy $\policy_{\theta}$ to acquire prior knowledge that guides generation of valid molecules.
The generator policy $\policy_{\theta}$, parameterized by a DNN with learned weights $\theta$, is defined as the product of probability distributions: 
$\policy_{\theta}\paren{\mathbf{y}|\mathbf{x}}=\prod_{t=1}^{|\mathbf{y}|} \policy_{\theta}\paren{y_t|\mathbf{x},\mathbf{y}_{<t}}$, where  $\policy_{\theta}\paren{\cdot|\mathbf{x},\mathbf{y}_{<t}}$ is a distribution, $\mathbf{y}_{<1}=y_0:=\text{[BOS]}$, and $\mathbf{x}$ is an input sequence.

Recall that the decoding process in text generation seeks to identify the most likely hypothesis from all potential candidates by solving the optimization problem:

\begin{equation}
\mathbf{y}^{\star}=\argmax_{\mathbf{y}\in \hypothesis_{\horizon}} \log \policy_{\theta}\paren{\mathbf{y}|\mathbf{x}}. 
\end{equation}

\fix{To estimate the expected reward for a partial molecule, we employ beam search to 
navigate the exponentially vast search space to form a complete valid molecule}.
Beam search (breadth-first search with breadth limited to $b\in \mathbb{R}^{+}$: the `beam')
can be expressed as the following recursion:

\begin{align}\label{eq:bs}
    \hypothesis_0&=\curlybracket{\text{[BOS]}},\\
    \hypothesis_t&=\argmax_{\hypothesis'\subseteq\mathcal{D}_t,\,|\hypothesis'|=\fix{b}} \log {\policy_{\theta}}\paren{\hypothesis'|\mathbf{x}},
\end{align}
where  
$\mathcal{D}_t=\curlybracket{\mathbf{y}_{t-1}\circ y ~|~ \mathbf{y}_{t-1}\in {\mathcal{V}} \text{~and~} \mathbf{y}_{t-1} \in \hypothesis_{t-1}}$ 
is the candidate set.
Because this method focuses on maximizing likelihood without concern for any specific metric of interest, it is not readily adaptable to optimize objectives that differ 
from those in its training set, as encoded in $\policy_{\theta}$.
Therefore, we cannot directly employ these generation algorithms to create molecules that achieve \fix{a different objective} like a significant docking score for a specific target site.

\textbf{MCTS.}
We employ a MCTS planning algorithm to identify the optimal policy in a MDP. 
This algorithm employs a tree structure for its search, where each node represents a state $\state$ and each edge an action $\action$. For every node $\state$, the algorithm tracks the visit count $\Num\paren{\state}$.
Additionally, $\Num\paren{\state,\action}$ represents the frequency with which action $\action$ has been selected from state $\state$ during the tree's construction.
The algorithm maintains an action value function, $\QFunc\paren{\state,\action}$, representing the best reward received by beginning in state $\state$ and executing action $\action$. 

UCT~\citep{kocsis2006bandit} (Upper Confidence bounds applied to Trees) adopts the UCB~\citep{auer2002finite} algorithm from Multi-Arm bandit for node selection:
\begin{equation}\label{eq:UCB}
 \text{UCB}=\QFunc\paren{\state,\action}+c_p\cdot\sqrt{\frac{\log\paren{\Num\paren{\state}}}{\Num\paren{\state,\action}}}, 
\end{equation}
so as to balance between exploiting states known to be advantageous and exploring those less visited.
Value-guided MCTS (V-MCTS)~\citep{silver2017mastering,zhang2023planning} improves UCT's sample efficiency by integrating both a policy $\policy$ and a value network $\VFunc$,
yielding P-UCT:
\begin{align}\label{eq:P-UCT}
        \text{P-UCT}\paren{\state}&=\argmax_{\action\in \actionSpace} \text{P-UCB}\paren{\state,\action},
\end{align}
where:
\begin{align}
    \text{P-UCB}\paren{{\state,\action}}&={\QFunc\paren{\state,\action}+c_{{p}}\cdot 
    \policy_{\tau}\paren{\action|\state}\cdot \frac{\sqrt{\log\paren{\Num\paren{\state}}}}{1+\Num\paren{\state,\action}}},\notag 
\end{align}
$c_{{p}}$ is a tunable constant, and $\tau$ represents a temperature parameter that modifies the policy  $\policy_{\tau}\paren{\action|\state}=\frac{\policy\paren{\action|\state}^{\frac{1}{\tau}}}{\sum_b \policy\paren{b|\state}^{\frac{1}{\tau}}}$. 
PG-TD~\citep{zhang2023planning}, a state-of-the-art implementation of V-MCTS, uses a pre-trained Transformer as the policy network $\policy$ and applies it to code generation.

\textbf{Drug generation process.} 
Following \citet{liu23Drug}, we formalize the drug discovery problem within the context of Markov decision processes (MDP). 
Our goal is to train a generative policy $\policy_{\theta}$ 
to generate a high-quality sequence denoted as $Y_{1:\horizon}=\paren{y_1,\ldots,y_t,\ldots,y_{\horizon}},y_t\in \mathcal{V}$.
At each time step $t$, the state $\state_{t-1}$ includes the tokens generated so far, $\paren{y_1,\ldots,y_{t-1}}$, and the action $\action$ corresponds to choosing the subsequent token $y_t$.
We want the generative policy $\policy_{\theta}$, when starting from an initial state $Y_1$, to maximize the expected final reward for a complete sequence:
\begin{equation}\label{eq:drug_generation_objective}
J\paren{\theta}=\expctover{Y_1 \sim \fix{\stateDist_0}}{\reward_{\horizon}|\theta},
\end{equation}
where $\reward_{\horizon}$ is the reward calculated for a %
generated sequence.

\textbf{Limitation of previous works:} 
Conventional MCTS faces computational challenges when searching for high-quality molecules. 
Relying solely on pre-trained LLM decoding for drug discovery results in inconsistencies in generating valid and high-quality molecules and lacks adaptability for different objectives. 
The current state-of-the-art, PG-TD, combines a Transformer with MCTS in a simplistic manner that does not account adequately in decision making for the policy's (lack of) certainty. 
In so doing, it prioritizes exploitation over exploration.
A second deficiency of PG-TD is that it can apply only a single reward objective. 
In this work, we introduce \algname, an entropy-reinforced planning approach for the Transformer decoder that addresses the limitations of PG-TD and other previous efforts by enhancing the exploration and exploitation tradeoff and permitting multiple objective improvement in the generation process.

\begin{figure*}[ht!]
    \begin{subfigure}{1\textwidth}
        \centering
        \includegraphics[%
        width=14cm, 
        clip={0,0,0,0}]{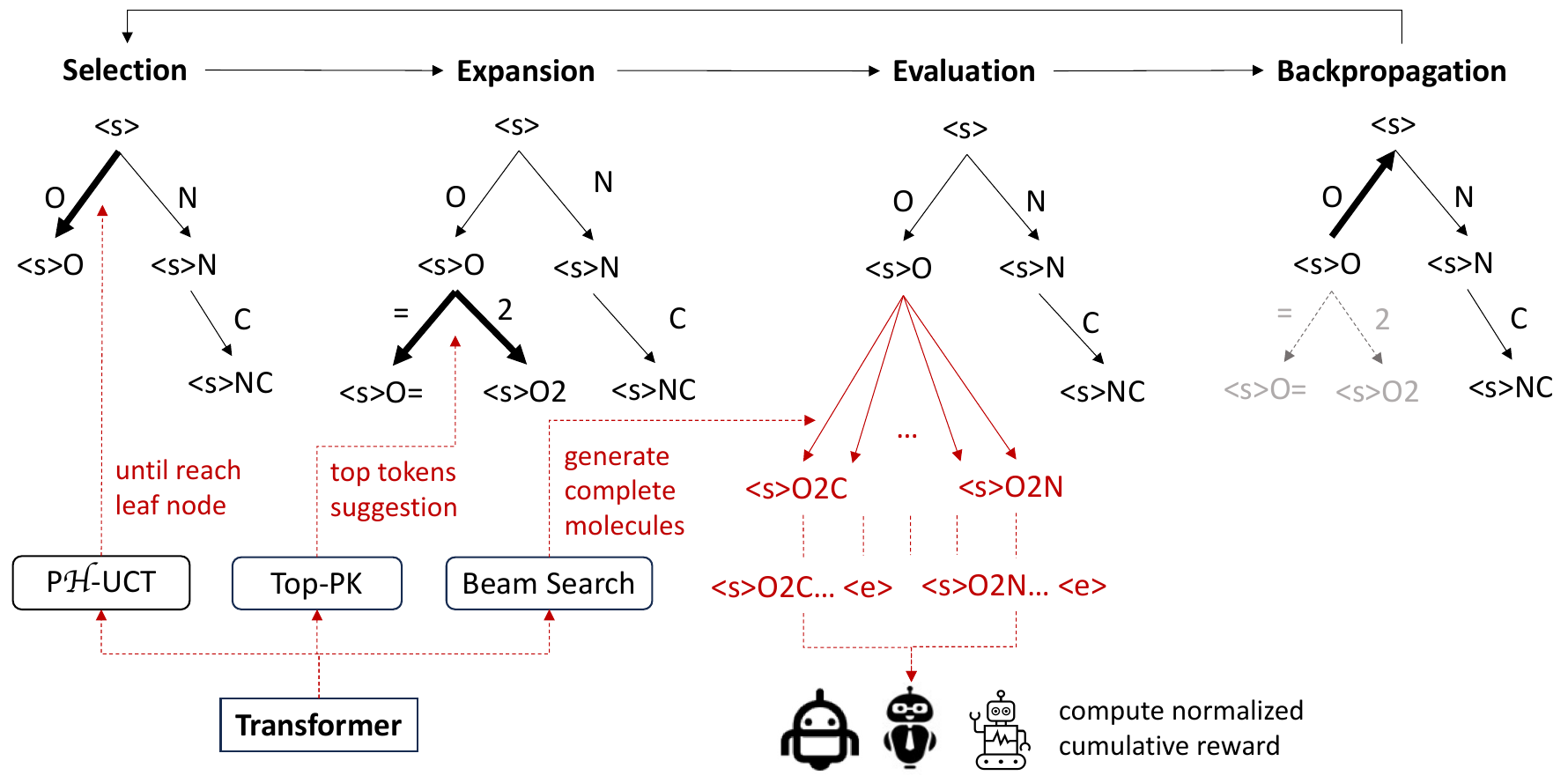}
    \end{subfigure}
    \caption{
Illustration of the application of the \algname algorithm in the Transformer's process for generating molecules. Here, <s> denotes the start token [BOS], and <e> signifies the end token [EOS]. The parts highlighted in {\color{Maroon}red} are from Transformer.
    }
\end{figure*}

\section{The \algname Algorithm}

Our Entropy-Reinforced Planning for Transformer Decoding (ECP) algorithm enhances Transformer decoding with an \emph{e}-steps entropy-reinforced MCTS planner. 
ECT enhances sample efficiency by using the Transformer's \fix{\BS, \TOPP, and \TOPK} functions to narrow the molecule search space. It also leverages the controlled generation capabilities of MCTS and improves exploration and exploitation through the use of \emph{e}-step forward entropy measurement.

We present the pseudocode of our algorithm in \algoref{alg:lops} and detail the entire process in the following. At each step of decoding, we allocate a fixed number of rollouts to construct a tree representing potential future trajectories. Each rollout involves four key steps:

\textbf{Selection.} 
We employ a new selection algorithm, $\PHUCT$, to decide which node to select. In this algorithm we enhance the bonus term to incorporate the probability of the next token as determined by the Transformer and its subsequent \emph{e}-step entropy. Thus, the tree search tends to select tokens that balance exploitation and exploration while reducing global uncertainty. The process involves recursively choosing child nodes based on the $\PHUCT$ formula, starting at the root and continuing until an unexplored leaf node $\state$ is reached:

\begin{align}\label{eq:p-uct-k}
    &\PHUCT\paren{\state_t,e}= \argmax_{\action\in \actionSpace} \PHUCB
    \paren{\paren{\state_t,\action},e},\\
    &\quad\quad\quad\text{where}~ \PHUCB\paren{\paren{\state_t,\action},e}= \QFunc\paren{\state_t,\action}\\
    &+c_{p} \frac{\sqrt{\log\Num\paren{\state_t}}}{1+\Num\paren{\state_t,\action}} \cdot \underbrace{\policy_{\tau}\paren{\action|\state_t} 
    \frac{1}{e} \sum_{i=1}^e {\mathcal{H}\paren{\policy_{\tau}\paren{\cdot|\state_{t+i}}}}}_{\text{Entropy-Reinforced Planning}}\notag,
\end{align}
$c_{p}$ is a tunable constant, $\tau$ is a temperature parameter applied to the policy $\policy_{\tau}\paren{\action|\state}=\frac{\policy\paren{\action|\state}^{\frac{1}{\tau}}}{\sum_b \policy\paren{b|\state}^{\frac{1}{\tau}}}$, and $\frac{1}{e} \sum_{i=1}^e {\mathcal{H}\paren{\policy_{\tau}\paren{\cdot|\state_{t+i}}}}$ is the \emph{e}-steps averaged entropy defined as follows:
\begin{align}
 \frac{1}{e} \sum_{i=1}^e {\mathcal{H}\paren{\policy_{\tau}\paren{\cdot|\state_{t+i}}}}=& \frac{1}{e} \sum_{i=1}^e \sum_{\fix{\action\in \mathcal{V}^{\star}}}\policy_{\tau}\paren{\action|\state_{t+i}}\\
 &\cdot\log\paren{\policy_{\tau}\paren{\action|\state_{t+i}}},\notag
\end{align}
where $\mathcal{V}^{\star}$ represents the candidate token space conditioned on $\state_{t+i}$.
Intuitively, the function $\PHUCT\paren{\state_t,e}$ is more likely to select an action $\action$ if 1) $\QFunc\paren{\state_t,\action_t}$ is high, indicating the discovery of a high reward molecule with prefix $\state_{t+1}$; or 2) $\policy_{\tau}\paren{\action|\state_t}$ is high, suggesting that the Transformer predicts $\action$ as a highly probable next token;  or 3) $\Num\paren{\state_t}$ is large while $\Num\paren{\state_t,\action}$ is small, implying that $\state_{t+1}$ has not been adequately explored.
Additionally, 4) the selection is favored when the entropy of the subsequential tree is large in expectation,
suggesting high uncertainty in its subsequence after performing action $\action$.

\textbf{Expansion.} When $\PHUCT$  selects a node $\state_{t+1}$
which has no leaf nodes, we need to identify potential next tokens and incorporate the corresponding next states as new child nodes.
Unlike standard MCTS, which may randomly sample a token, potentially leading to an invalid molecule, we employ a mixture of the $\TOPP$  and $\TOPK$ functions: 
\begin{align}\label{eq:toppk}
    &\text{\TOPPK}\paren{\mathbf{y}_{<i}, p,k}= \actionSpace_{\mathbf{y}_{<i}},\\ \notag
    &\text{~~~~~where~} \actionSpace_{\mathbf{y}_{<i}}=\fix{\curlybracket{y_1,\ldots,y_i,\ldots,y_{j}}, y_i\in \mathcal{V}}, 
    \\ \notag
    &\text{~~~~~~} j= \min \curlybracket{\argmin_{j'}\sum_{l=1}^{j'}\policy_{\theta}\paren{y_l|\mathbf{y}_{<i}} \geq p, k},\\ \notag
    &\text{~~~~~~} \text{and } \policy_{\theta}\paren{y_g|\mathbf{y}_{<i}}>\policy_{\theta}\paren{y_h|\mathbf{y}_{<i}}, \forall g<h. \notag
\end{align}
For a given state $\state_{t+1}$, \text{\TOPPK} $\paren{\state_{t+1}, p, k}$ \eqref{eq:toppk} of the pretrained Transformer retrieves the at most $k$ probable subsequent tokens $\actionSpace_{\mathbf{y}_{<i}}$ \fix{with a maximum accumulated probability $p$} based on prior experience, where $k$ is the maximum number of children. The \fix{at most} $k$ next states, formed by appending each of these suggested tokens from the Transformer to the current state, are then added to the children list of the current node. \fix{After the tree is thus expanded,}
we then need to evaluate the selected node $\state_t$.

\textbf{Evaluation.} The selected node $\state_t$ might still be an incomplete molecule, but our reward function only calculates the reward for a complete molecule. To evaluate a partial molecule,
we employ the $\BS\paren{\state_t,b}$ \eqref{eq:bs} function from Transformer to generate complete molecules $\state_{\horizon}$ with a prefix of partial molecule $\state_t$ and beam size $b$. 
Generated molecules are evaluated with the \fix{reward function},
and the result is used as the value of node $\state_t$.

\textbf{Backpropagate.} 
The reward \fix{of the completed molecule}, denoted $\reward_{{{\horizon}}}$, is recursively propagated up the tree from the initially selected node until it reaches the root. During this tree traversal, each value $\QFunc\paren{\state_i,\action_i}|_{0 \leq i<t}$ encountered is updated with the newly obtained value $\reward_{\horizon}$, via the formula 
\begin{equation}\label{eq:backprobagate}
\QFunc\paren{\state_i,\action_i}\leftarrow \max \curlybracket{\QFunc\paren{\state_i,\action_i},\reward_{\horizon}},
\end{equation}
where $0\leq i<t$, and $\state_i\circ\action_i=\state_{i+1}$.

\begin{figure}[ht]
    \begin{subfigure}{1\textwidth}
        \includegraphics[height=6cm, 
        clip={0,0,0,0}]{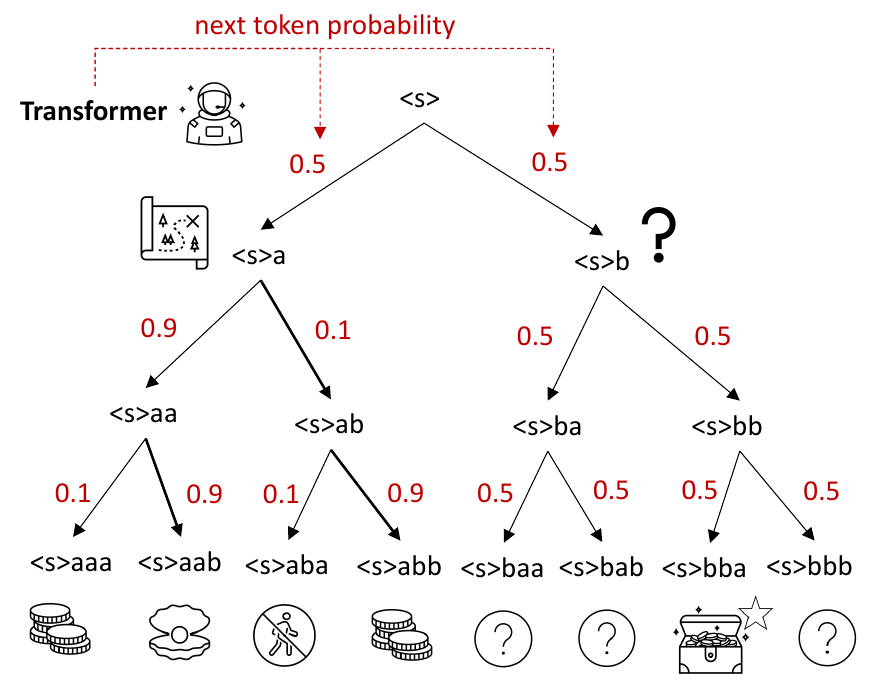}
    \end{subfigure}
    \caption{An environment with action space $\actionSpace = {\text{left}, \text{right}}$, in which each node (state) is connected by only two edges (actions), and each edge is associated with a probability of being sampled, as determined by the pretrained Transformer decoder $\policy_{\theta}$. The {\color{Maroon}red} values are inferred by the Transformer.
    }
  \label{fig:entropy_example}
\end{figure}

\textbf{Illustration of how \emph{e}-step entropy makes the difference.}
\figref{fig:entropy_example} assumes an environment with the initial state (partial molecule) $\state_0=\text{<s>}$ and an action space defined as $\actionSpace=\curlybracket{\text{left},\text{right}}$.
\fix{Given a pretrained Transfomer decoder $\policy_{\theta}$,}
if we consider only the initial $\state_0$, then $\policy_{\theta}$ presents equal probability for both actions: $\policy_{\theta}\paren{\text{left}|\mathbf{x}}=\policy_{\theta}\paren{\text{right}|\mathbf{x}}$, where $\mathbf{x}=\text{<s>}$.
However, if we look a few steps ahead,  the subsequent tree under the right action presents much higher uncertainty that does that under the left action, \fix{indicating that the Transformer has less experience on the right-hand branch.}
Thus, in this scenario, we need to reduce the uncertainty of $\policy_{\theta}$ via greater exploration of the right tree, if we are not to miss the hidden high-reward state.

\textbf{Multi-critic \fix{normalized reward}.} We have an ensemble array of critics as follows:
\begin{equation*}
\mathbf{C}= \bracket{\critic^{\text{Druglikeness}}, \critic^{\text{Solubility}}, \critic^{\text{Synthesizability}}, \critic^{\text{Docking}}},  
\end{equation*}
where each critic $\critic: Y_{1:\horizon} \rightarrow \mathbb{R}$ \fix{represents a unique quality evaluator of a pharmaceutical property}.
Here we reference the multiple critic from \citet{liu23Drug}, and designed a normalized reward function to align the drug optimization with multiple objectives as follows, 
\begin{equation}\label{eq:normalized_score}
\RFunc^{\text{sum}}_{\text{norm}}{\paren{\state_{\horizon}}}=
\sum_{i=0}^{|\mathbf{C}|-1}\text{Norm}\paren{\critic_i{\paren{\state_{\horizon}}}},    
\end{equation}
where $\critic: Y_{1:\horizon} \rightarrow \mathbb{R},\text{and }\critic_i\in \mathbf{\critic}$.
{We use Norm to normalize different attributes {onto the} same scale.} {{Here, we define Norm as min-max normalization to scale the attributes onto the range [0, \fix{1}].}}

\begin{algorithm}[H]
\caption{Entropy-Reinforced Planning (\algname) 
}
\label{alg:lops}
\begin{algorithmic}[1]
\Require $root$: $\state_0$,
$c_{\text{p}}$: exploration parameter\\
\RNum: rollout number\\
$p$: maximum cumulative probability\\
$k$: maximum number of node's children\\
$b$: beam number\\
$e$: forward entropy steps\\
${cache}=\text{DICTIONARY}\paren{}$.

\For{$i\leftarrow 1,2,...,\RNum$}
       \State $node$ $\leftarrow$ $root$. 
       \LineComment{/* Selection */}
       \While {|$node.children$| > 0 }
       \State $node$ $\leftarrow \PHUCT \paren{{node.child}, e}$ in Eq.~\eqref{eq:p-uct-k}
       \EndWhile
       
       \LineComment{/* Expansion */}
       \State $[{next\_tokens}] \leftarrow \text{\TOPPK}\paren{node, p, k}$ Eq.~\eqref{eq:toppk}
       \For {$next\_token \in  [next\_tokens]$}
       \State $next\_state \leftarrow \text{CONCAT}\paren{node, next\_token}$
       \State Create a node ${new\_node}$ for ${next\_state}$
       \State Add ${new\_node}$ to the children of $node$
       \EndFor
       \LineComment{/* Evaluation */}
       \State $ \curlybracket{mols} \leftarrow \BS \paren{node,b}$

       \For{$mol \in \curlybracket{mols}$ and {$mol$} $\notin$ $cache$.keys()}
       
       \State $r\leftarrow \text{GET-REWARD}\paren{mol}$ in \eqref{eq:normalized_score}
       \State ${cache} \bracket{mol} = r$ 
       \EndFor
       \State $\reward_{\horizon}=\max \curlybracket{cache[mol]}, mol\in \curlybracket{mols}$
       \LineComment{/* Backpropagation */}
       \State Update the values of $node$ with $\reward_{\horizon}$
       \State Update its ancestors with $\reward_{\horizon}$ by Eq. \eqref{eq:backprobagate}
\EndFor
\Return Top reward molecules in ${cache}$
\end{algorithmic}
\end{algorithm}

\begin{table*}[t!]
\setlength{\tabcolsep}{4pt}
   \centering
    {\small
    \scalebox{0.75}{
    \begin{tabular}{l l c c c c c c c c }
        \toprule
        \textbf{Target} %
        & \textbf{Algorithm}
        & {\makecell[c]{Best\\ Norm Reward~$\uparrow$}}
        & {\makecell[c]{Avg Valid\\ Norm Reward~$\uparrow$}}
        & {\makecell[c]{Avg Top 10\%\\ Norm Reward~$\uparrow$}}
        & {\makecell[c]{Unique\\ Valid Molecule~$\uparrow$}}
        & {\makecell[c]{Docking ~$\downarrow$}}
        & {\makecell[c]{Druglikeliness ~$\uparrow$}}
        & {\makecell[c]{Synthesizability ~$\downarrow$}}
        & {\makecell[c]{Solubility ~$\uparrow$}}
        \\
        \midrule
        \makecell[l]{\textbf{3CLPro}} %
        &  \textbf{\makecell[l]{Beam Search}}
        &  \makecell[l]{2.53 \ (-10.99\%) }
        &  \makecell[l]{2.38 \ (-0.75\%)} 
        &  \makecell[l]{2.53 \ (-7.40\%)} 
        &  \makecell[l]{16 \ (-98.63\%)}
        &  \makecell[l]{-8.67 \ (6.30\%)}
        &  \makecell[l]{0.86 \ (26.91\%)}
        &  \makecell[l]{3.35 \ (-59.18\%)}
        &  \makecell[l]{3.01 \ (-31.87\%)}
        \\
        (PDBID:
        &  \textbf{\makecell[l]{Sampling}}
        &  \makecell[l]{2.85 \ (0\%) }
        &  \makecell[l]{2.39 \ (0\%) }
        &  \makecell[l]{2.74 \ (0\%)}
        &  \makecell[l]{1609 \ (0\%)}
        &  \makecell[l]{-8.15 \ (0\%)}
        &  \makecell[l]{0.68 \ (0\%)}
        &  \makecell[l]{2.11 \ (0\%)}
        &  \makecell[l]{4.42 \ (0\%)}
        \\
       \ 7BQY)
        &  \textbf{\makecell[l]{\fix{UCT}}}
        &  \makecell[l]{3.10 \ (8.86\%) }
        &  \makecell[l]{2.25 \ (-5.83\%)} 
        &  \makecell[l]{2.65 \ (-3.30\%)} 
        &  \makecell[l]{\textbf{3013 \ (157.96\%)}}
        &  \makecell[l]{-8.45 \ (3.67\%)}
        &  \makecell[l]{\textbf{0.89 \ (31.34\%)}}
        &  \makecell[l]{2.02 \ (4.17\%)}
        &  \makecell[l]{4.39 \ (-0.84\%)}
        \\
        \textbf{ }
        &  \textbf{\makecell[l]{PG-TD}}
        &  \makecell[l]{3.20 \ (12.50\%)} 
        &  \makecell[l]{2.59 \ (8.05\%)} 
        &  \makecell[l]{2.99 \ (9.38\%)} 
        &  \makecell[l]{2549 \ (118.24\%)} 
        &  \makecell[l]{-8.37 \ (2.70\%)} 
        &  \makecell[l]{0.88 \ (30.30\%)} 
        &  \makecell[l]{1.79 \ (15.15\%)} 
        &  \makecell[l]{\textbf{4.79 \ (8.37\%)}}
        \\
        \textbf{ }
        &  \textbf{\makecell[l]{\algname(Ours)}}
        &  \makecell[l]{\textbf{3.24  \ (13.71\%)}}
        &  \makecell[l]{\textbf{2.62 \ (9.42\%)}} 
        &  \makecell[l]{\textbf{3.07 \ (12.25\%)}} 
        &  \makecell[l]{2575 \ (120.46\%)}
        &  \makecell[l]{\textbf{-9.13 \ (11.98\%)}}
        &  \makecell[l]{\textbf{0.89 \ (31.00\%)}}
        &  \makecell[l]{\textbf{1.71 \ (18.78\%)}}
        &  \makecell[l]{4.44 \ (0.50\%)}
        \\
        \bottomrule
        \textbf{RTCB}
        &  \textbf{\makecell[l]{Beam Search}}
        &  \makecell[l]{2.56 \ (-17.46\%)}
        &  \makecell[l]{2.39 \ (-3.35\%)} 
        &  \makecell[l]{\fix{2.39} \ (-17.02\%)}
        &  \makecell[l]{8 \ (-99.32\%)}
        &  \makecell[l]{-7.11 \ (-30.85\%)}
        &  \makecell[l]{0.82 \ (-3.55\%)}
        &  \makecell[l]{3.02 \ (-55.02\%)}
        &  \makecell[l]{3.55 \ (1.68\%)}
        \\
        (PDBID:
        &  \textbf{\makecell[l]{Sampling}}
        &  \makecell[l]{3.10 \ (0\%)}
        &  \makecell[l]{2.47 \ (0\%)}
        &  \makecell[l]{2.88 \ (0\%)} 
        &  \makecell[l]{1168 \ (0\%)}
        &  \makecell[l]{\textbf{-10.28 \ (0\%)}}
        &  \makecell[l]{0.85 \ (0\%)}
        &  \makecell[l]{1.95 \ (0\%)}
        &  \makecell[l]{3.49 \ (0\%)}

        \\
        \ 4DWQ)
        &  \textbf{\makecell[l]{\fix{UCT}}}
        &  \makecell[l]{2.97  \ (-4.12\%)}
        &  \makecell[l]{2.28 \ (-7.61\%)} 
        &  \makecell[l]{2.73 \ (-5.36\%)} 
        &  \makecell[l]{\textbf{1491 \ (27.65\%)}}
        &  \makecell[l]{-8.62 \ (-16.22\%)}
        &  \makecell[l]{0.83 \ (-2.77\%)}
        &  \makecell[l]{1.89 \ (3.18\%)}
        &  \makecell[l]{3.41 \ (-2.46\%)}
        \\
        \textbf{ }
        &  \textbf{\makecell[l]{PG-TD}}
        &  \makecell[l]{3.10 \ (0.05\%)} 
        &  \makecell[l]{2.53 \ (2.26\%)} 
        &  \makecell[l]{2.93 \ (1.63\%)} 
        &  \makecell[l]{1266 \ (8.39\%)} 
        &  \makecell[l]{-9.22 \ (-10.30\%)} 
        &  \makecell[l]{0.84 \ (-1.82\%)} 
        &  \makecell[l]{1.72 \ (11.88\%)} 
        &  \makecell[l]{3.67 \ (5.03\%)} 
        \\
        \textbf{ }
        &  \textbf{\makecell[l]{\algname(Ours)}}
        &  \makecell[l]{\textbf{3.24 \ (4.52\%)}}
        &  \makecell[l]{\textbf{2.61 \ (5.61\%)}}
        &  \makecell[l]{\textbf{3.07 \ (6.68\%)}}
        &  \makecell[l]{1343 \ (14.98\%)}
        &  \makecell[l]{-8.81 \ (-14.34\%)}
        &  \makecell[l]{\textbf{0.93 \ (9.40\%)}}
        &  \makecell[l]{\textbf{1.56 \ (20.01\%)}}
        &  \makecell[l]{\textbf{3.87 \ (10.92\%)}}
        \\
        \bottomrule
        \\

    \end{tabular}}}
        \caption{
        {\textbf{Main results.} A comparison of four baselines \{\text{Beam Search}, \text{Sampling}, \text{UCT}, \text{PG-TD}\} and \algname on multiple objectives 
        based on 3CLPro and RTCB datasets and pretrained Transformer model. 
        Boldface denotes best.
        The percentage of improvement over the \fix{Sampling} baseline is enclosed in parentheses.
        All experiments are conducted with 256 rollouts.
        }
        \label{exp:main_result}
        }
\end{table*}

\section{Experiments}

We describe our experimental setup and present results.

\subsection{Drug discovery}

\subsubsection{Experimental configuration}

\textbf{The language model.} We train three GPT-2-like Transformers on the task of causal language modeling, as follows:
    
    The \textit{pretrained model} is a 124M GPT2 model trained using a BPE tokenizer~\citep{bostrom2020byte} on a diverse dataset of 10.7 million SMILES strings of drug-like molecules 
    randomly sampled from the ZINC database \citep{irwin2005zinc}.
    
    The \textit{biased model} is fine-tuned based on the pretrained model
    with a distinct objective focused solely on the docking score optimization, using a biased dataset that only contains molecules with docking scores in the range \fix{[-6, -14]}.
    \fix{Docking scores are numerical values generated by computational docking simulations, predicting how well a small molecule, such as a potential drug, fits into the binding site of a target protein. A higher score represents better strength and stability of the interaction.}
    
    The \textit{RL fine-tuned model} is fine-tuned based on a pretrained model, using the same objective as \algname, with the multi-critic normalized reward function \eqref{eq:normalized_score}. We fine-tune the pretrained model by using an RL approach~\citep{liu23Drug} focusing on the top 10 \fix{sampled} molecules optimization. 
    
\paragraph{Baselines.} We compare \algname with four baselines:
1) \emph{Beam search};
2) \emph{Sampling}, in which the Transformer's top-k sampling algorithm is employed to sample a token from the top-k most likely tokens at each step to generate a set of molecules;
3) \emph{UCT}, which adopts the UCB algorithm for node selection; and
4) \emph{PG-TD},  the current state-of-the-art algorithm for integrating MCTS and Transformer by applying P-UCT~\eqref{eq:P-UCT}.
Both UCT and PG-TD are tailored to use the reward function \eqref{eq:normalized_score}.

\paragraph{Dataset.}
We employ, from the most recent Cancer and COVID dataset of \citet{liu23Drug}, 1 million compounds from the ZINC15 dataset docked to the 3CLPro~(PDB ID: 7BQY) protein associated with SARS-CoV-2 and the RTCB (PDB ID: 4DWQ) human cancer protein.

\begin{figure*}[t!]
    \begin{subfigure}{.24\textwidth}
        \centering
        \includegraphics[%
        width=3.4cm,  clip={0,0,0,0}]{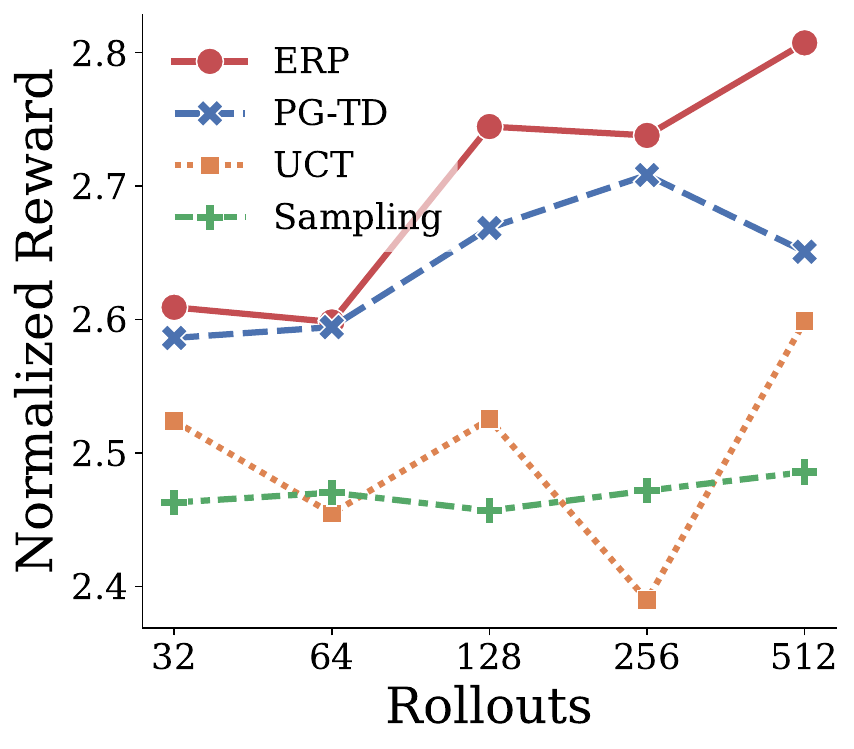}
        \caption{\fix{Pretrained} model (RTCB)}\label{fig:results:pretrain:cancer}
    \end{subfigure}\hfil
    \begin{subfigure}{.24\textwidth}
        \centering
        \includegraphics[%
        width=3.4cm,  clip={0,0,0,0}]{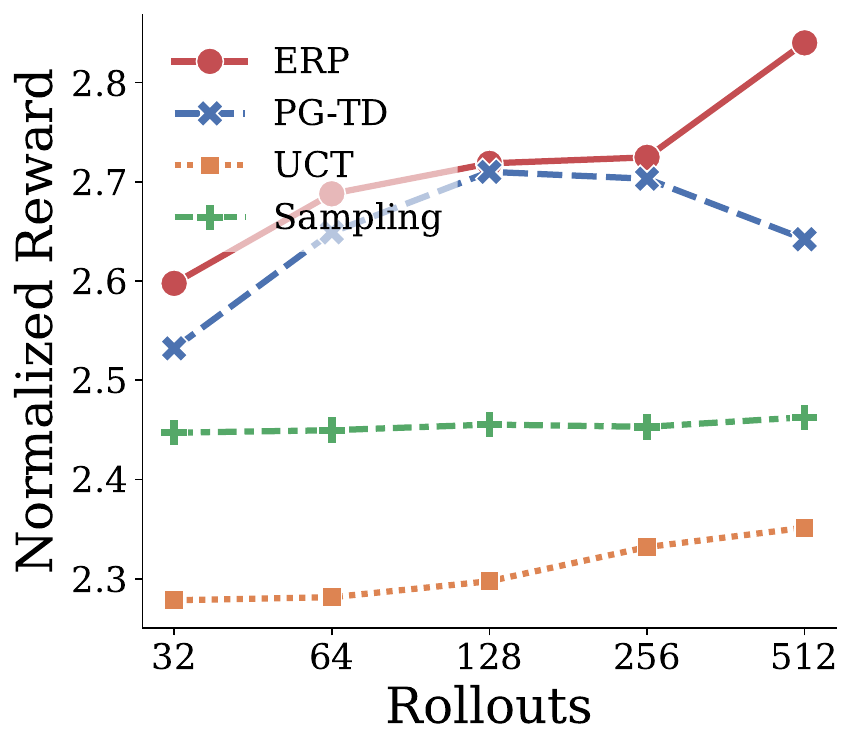}
        \caption{\fix{Pretrained} model (3CLPro)}\label{fig:results:pretrain:covid}
    \end{subfigure}\hfil
    \begin{subfigure}{.24\textwidth}
        \centering
        \includegraphics[%
        width=3.6cm,  clip={0,0,0,0}]{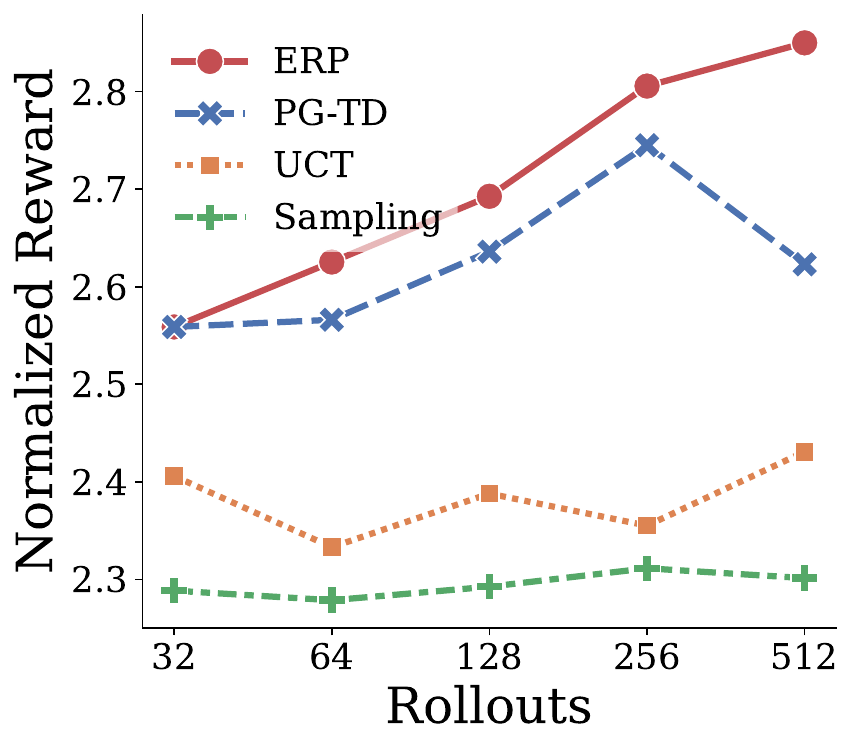}
        \caption{Biased model (RTCB)}\label{fig:results:biased:cancer}
    \end{subfigure}\hfil
    \begin{subfigure}{.24\textwidth}
        \centering
        \includegraphics[%
        width=3.4cm,  clip={0,0,0,0}]{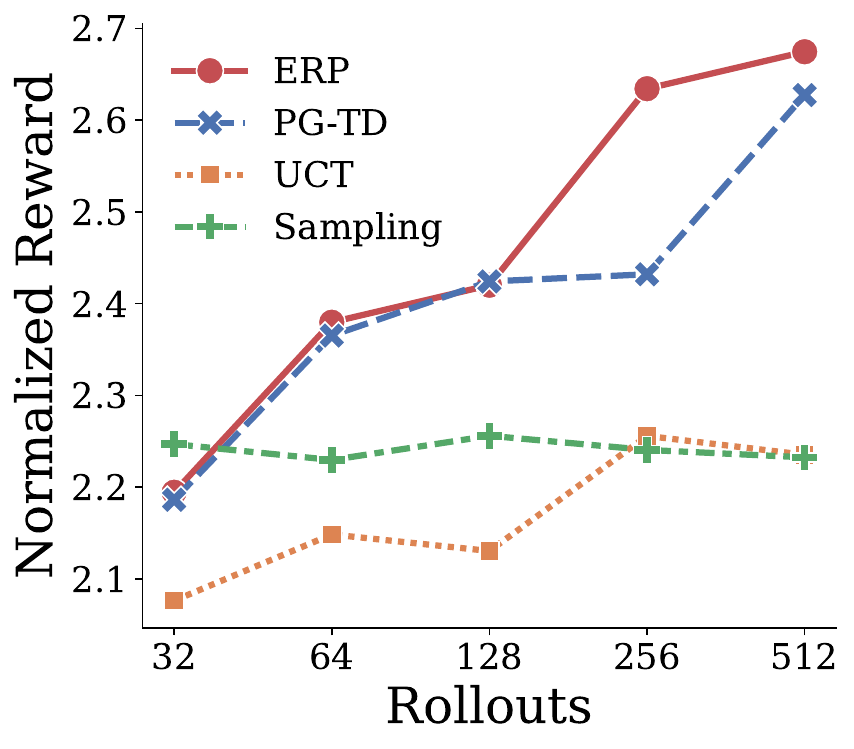}
        \caption{Biased model (3CLPro)}\label{fig:results:biased:covid}
    \end{subfigure}\hfil

    \begin{subfigure}{.24\textwidth}
        \centering
        \includegraphics[%
        width=3.4cm,  clip={0,0,0,0}]{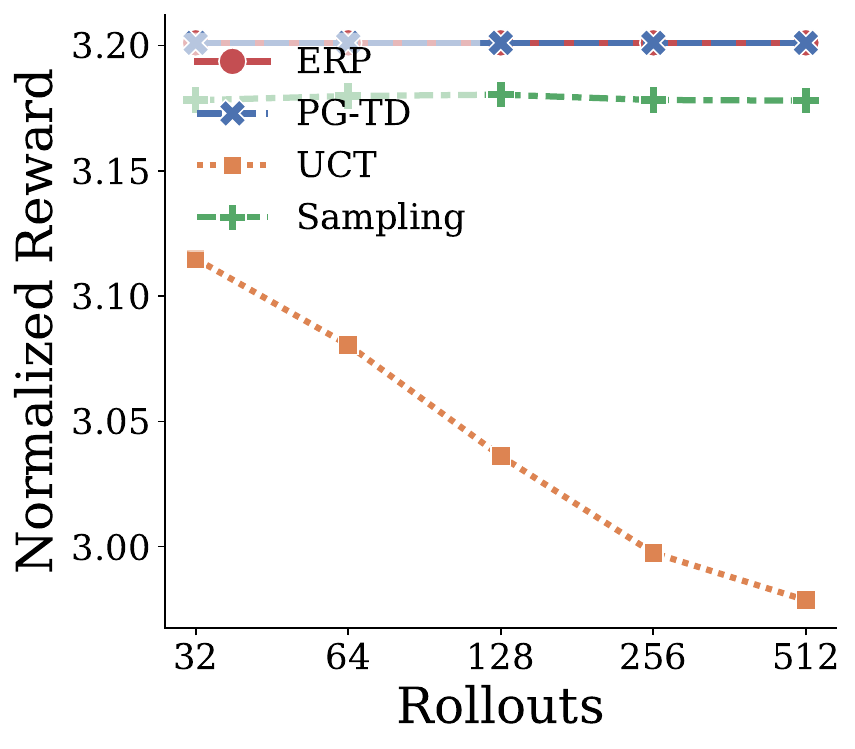}
        \caption{RL fine-tuned mod.\  (3CLPro)}\label{fig:results:finetune:covid}
    \end{subfigure}
    \begin{subfigure}{.24\textwidth}
        \centering
        \includegraphics[%
        width=3.4cm,  clip={0,0,0,0}]{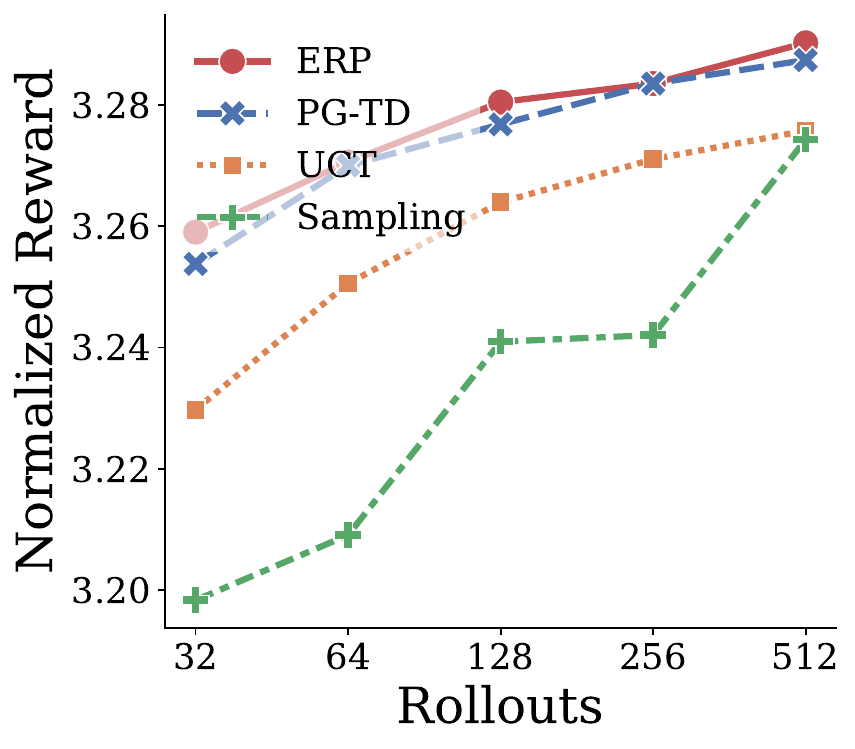}
        \caption{Top 10 leading compounds}\label{fig:results:finetune:covid10}
    \end{subfigure}\hfil
    \begin{subfigure}{.24\textwidth}
        \centering
        \includegraphics[%
        width=3.4cm,  clip={0,0,0,0}]{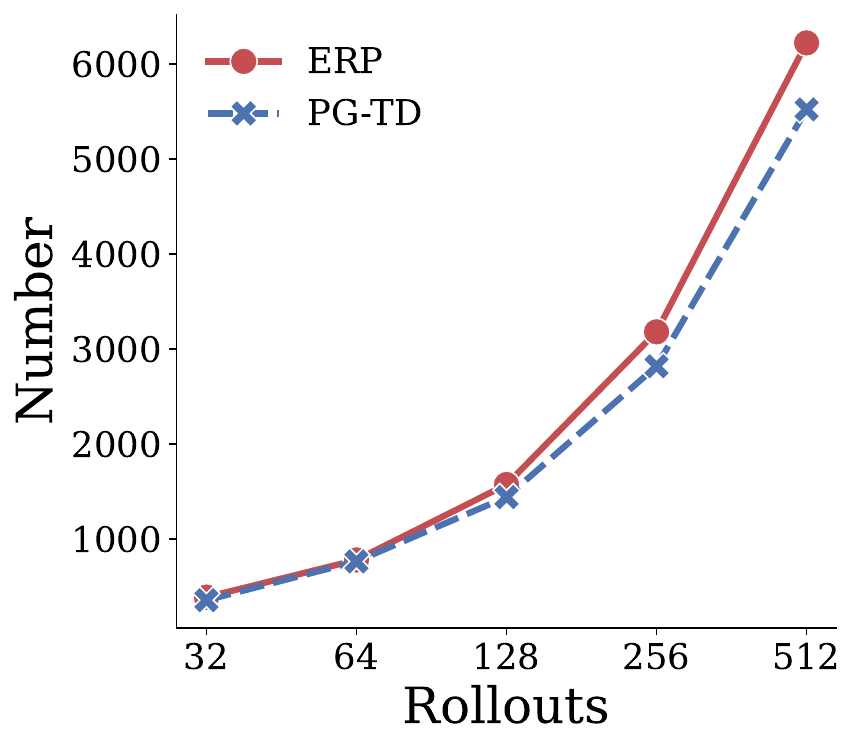}
        \caption{Num. unique valid molecules
        }\label{fig:results:rollout_num}
    \end{subfigure}\hfil
    \begin{subfigure}{.24\textwidth}
        \centering
        \includegraphics[%
        width=3.4cm,  clip={0,0,0,0}]{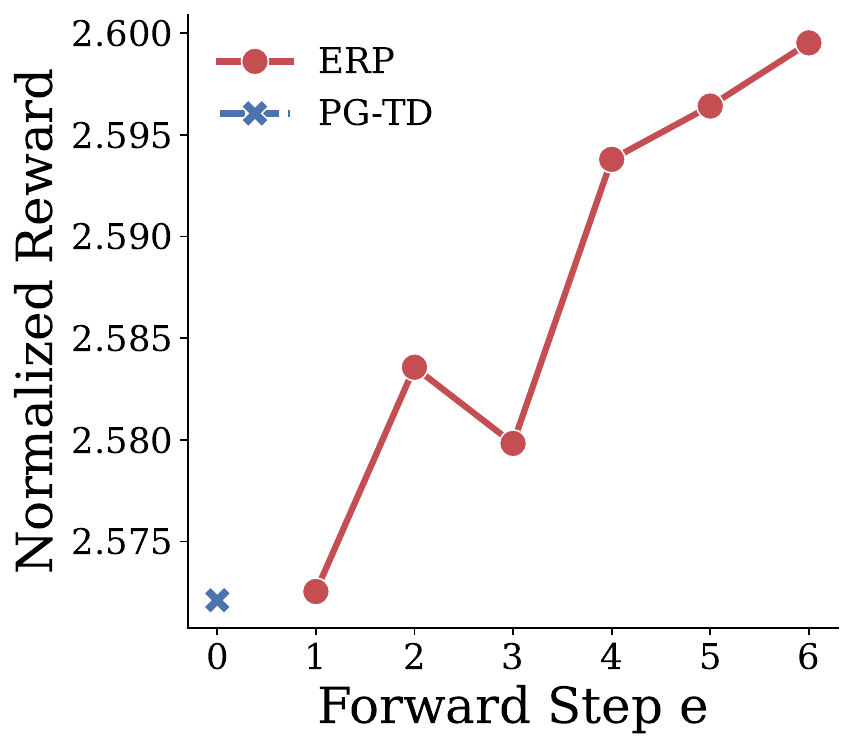}
        \caption{ERP Step \(e\) (3CLPro)
        }\label{fig:results:e_steps}
    \end{subfigure}\hfil
    \caption{\small \textbf{Ablation studies.} (a)(b)(c)(d)(e) Normalized rewards averaged among valid molecules for different LLMs 
    and algorithms.
    { \algname is our model.  PG-TD \citep{zhang2023planning} is the previous state-of-the-art method as described by Eq.~(\ref{eq:P-UCT}), and UCT by Eq~(\ref{eq:UCB}).  We also do random sampling from the LM as the Sampling baseline.}
    (f) Filtered the top 10 leading compounds from the molecules discovered in (e). 
    (g) \algname vs.\ PG-TD for number of unique valid molecules in \fix{3CLPro} dataset across different rollouts.
    (h) Effects of entropy step \(e\) of \algname. 
    }\label{fig:exp:ablation}
\end{figure*}

\paragraph{Critics and evaluation metric.} 
We evaluate seven key attributes for pharmaceutical drug discovery:
1) \emph{Best normalized reward} is the molecule with the top normalized reward;
2) \emph{Average valid normalized reward} is the average normalized reward of all valid molecules;
\fix{3) \emph{Average top 10\% normalized reward} is the average normalized reward of top 10\% valid molecule.}
4) \emph{Unique valid samples}; 
5) \emph{Druglikeness} assesses the probability of a molecule being a suitable drug candidate;
6) \emph{Solubility} evaluates a molecule's water-octanol partition coefficient (LogP), indicating how well it can dissolve in water;
7) \emph{Synthesizability} measures the synthesizability of a molecule, assigning a score of 1 for easy synthesis and a score of \fix{10} for difficult synthesis \fix{~\citep{ertl2009estimation}}; and
8) \emph{Docking score} (generated, for efficient calculation, with a surrogate docking model: see Appendix \ref{app:surrogate_model})
evaluates the potential of a drug to inhibit the target site.

\subsubsection{Results}

The \algname (\textcolor{Maroon}{red}) and baseline results in 
Table~\ref{exp:main_result} and \figref{fig:exp:ablation} show that \algname significantly outperforms all baselines, including the current state-of-the-art, PG-TD (in \textcolor{NavyBlue}{blue}), across various performance metrics. (For even more baselines of simpler models, see appendix \ref{app:sec:extendedbaselines}.)  Notably, \algname enhances the performance of different types of pretrained Transformers (also used as the sampling baseline in {\textcolor{ForestGreen}{green}}) as follows:

\paragraph{Pretrained LLM} serves as a general model. Figs~\ref{fig:results:pretrain:cancer} and \ref{fig:results:pretrain:covid} show that the Transformer-based planning algorithms, \algname and PG-TD, outperform the non-Transformer UCT significantly. This result suggests that the prior experience of the pretrained Transformer benefits both \algname and PG-TD by improving sampling efficiency. We note also that \algname surpasses PG-TD in both benchmarks. %

\paragraph{Biased LLMs,} trained on objectives different from those in this work, result in the Transformer Sampling baseline performing significantly worse than the pure MCTS UCT baseline. This result indicates that the biased prior experience of the Transformer guides molecule exploration into low-reward regions. However, Figs \ref{fig:results:biased:cancer} and \ref{fig:results:biased:covid} illustrate that \algname still outperforms the others in this situation, demonstrating the effectiveness of its controllable generation process. By adopting \algname's planning guidance, \fix{ the generation process} optimizes from the prior objective to the current one, showcasing its adaptability.

\paragraph{RL fine-tuned LLMs} 
present a challenging benchmark, involving optimization based on the top 10 sampled molecules until performance plateaus. Figs~\ref{fig:results:finetune:covid} and \ref{fig:results:finetune:covid10} demonstrate that \algname still enhances the capabilities of these optimized models. By implementing the entropy-reinforced planner, it continues to refine its performance through active exploration and exploitation of promising regions. Since the Transformer model is fine-tuned based solely on the top 10 samples, \figref{fig:results:finetune:covid} shows similar performance between \algname and PG-TD in terms of average valid molecules. Yet \figref{fig:results:finetune:covid10} reveals that \algname surpasses PG-TD in identifying the top 10 leading compounds.

\paragraph{Effectiveness of ERP.} Table~\ref{exp:main_result} and \figref{fig:exp:ablation} demonstrate the effectiveness of ERP. In Figs \ref{fig:results:pretrain:cancer}, \ref{fig:results:pretrain:covid}, and \ref{fig:results:biased:cancer}, PG-TD experienced a non-trivial performance drop when the number of rollouts increased from 256 to 512. This indicates that PG-TD began to discover mostly low-reward molecules, significantly lower than those discovered previously, thereby reducing the overall performance compared to that in the 256-rollout scenario. However, \algname enjoyed a monotonic performance improvement, finding molecules with even higher normalized values compared to previous rollouts. Moreover, \figref{fig:results:rollout_num} illustrates that \algname consistently discovered more unique samples than PG-TD. This metric assesses the algorithm's diversity and exploratory capability. These results indicates that PG-TD fails to conduct effective exploration as the number of rollouts grows, while \algname was capable of balancing exploration and exploitation, discovering higher-reward molecules as rollouts increase.  \fix{For more ablations, see Appendix~\ref{app:sec:moreablations}.}

\paragraph{Effect of \emph{e}-step forward in entropy measurement.}
We see in \figref{fig:results:e_steps} that performance improves with the number of forward steps for entropy-reinforced measurement. This improvement occurs because deeper analysis of the subtree enables better estimation of certainty for balancing between exploitation and exploration.

\subsection{\fix{Code generation}}
\begin{table*}[ht!]
\setlength{\tabcolsep}{10pt}
   \centering
    {\small
    \scalebox{0.93}{
    \begin{tabular}{c c c c c c c c c }
        \toprule
        \textbf{n@k}
        & {\makecell[l]{\textbf{Base}\\ \textbf{Model}}} %
        & \textbf{Algorithm}
        & {\makecell[c]{PR (\%)\\ Intro.}}
        & {\makecell[c]{PR (\%)\\ Inter.}}
        & {\makecell[c]{PR (\%)\\ comp.}}
        & {\makecell[c]{SA (\%)\\ Intro.}}
        & {\makecell[c]{SA (\%)\\ Inter.}}
        & {\makecell[c]{SA (\%)\\ comp.}}
        \\
        \toprule
        1@15
        &  \textbf{GPT-2}
        &  \textbf{\makecell[l]{PG-TD}}
        &  \makecell[l]{23.242}
        &  \makecell[l]{\textbf{12.481}}
        &  \makecell[l]{\textbf{19.055}}
        &  \makecell[l]{2.000}
        &  \makecell[l]{\textbf{3.433}}
        &  \makecell[l]{\textbf{8.600}}
        \\
        &
        &  \textbf{\makecell[l]{ERP (Ours)}}
        &  \makecell[l]{\textbf{23.955}}
        &  \makecell[l]{12.248}
        &  \makecell[l]{18.532} 
        &  \makecell[l]{2.000}
        &  \makecell[l]{3.267}
        &  \makecell[l]{7.900}
        \\
        1@15
        &  \textbf{GPT-Neo}
        &  \textbf{\makecell[l]{PG-TD}}
        &  \makecell[l]{\textbf{24.399}}
        &  \makecell[l]{12.301} 
        &  \makecell[l]{20.333} 
        &  \makecell[l]{\textbf{2.200}}
        &  \makecell[l]{3.433}
        &  \makecell[l]{9.200}
        \\
        &
        &  \textbf{\makecell[l]{ERP (Ours)}}
        &  \makecell[l]{24.384}
        &  \makecell[l]{\textbf{12.490}}
        &  \makecell[l]{\textbf{21.117}} 
        &  \makecell[l]{2.100}
        &  \makecell[l]{\textbf{3.467}}
        &  \makecell[l]{\textbf{10.100}}
        \\
        \midrule
        1@20
        &  \textbf{GPT-2}
        &  \textbf{\makecell[l]{PG-TD}}
        &  \makecell[l]{23.540}
        &  \makecell[l]{12.642}
        &  \makecell[l]{18.808}
        &  \makecell[l]{2.200}
        &  \makecell[l]{3.333}
        &  \makecell[l]{8.400}
        \\
        &
        &  \textbf{\makecell[l]{ERP (Ours)}}
        &  \makecell[l]{\textbf{25.204}}
        &  \makecell[l]{\textbf{13.277}}
        &  \makecell[l]{\textbf{20.174}} 
        &  \makecell[l]{\textbf{2.300}}
        &  \makecell[l]{\textbf{3.467}}
        &  \makecell[l]{\textbf{9.000}}
        \\
        1@20
        &  \textbf{GPT-Neo}
        &  \textbf{\makecell[l]{PG-TD}}
        &  \makecell[l]{24.525}
        &  \makecell[l]{12.392} 
        &  \makecell[l]{20.541} 
        &  \makecell[l]{2.200}
        &  \makecell[l]{3.333}
        &  \makecell[l]{9.400}
        \\
        &
        &  \textbf{\makecell[l]{ERP (Ours)}}
        &  \makecell[l]{\textbf{26.034}}
        &  \makecell[l]{\textbf{13.978}}
        &  \makecell[l]{\textbf{21.342}} 
        &  \makecell[l]{\textbf{2.400}}
        &  \makecell[l]{\textbf{3.933}}
        &  \makecell[l]{\textbf{10.200}}
        \\
        \midrule
        pass@15
        &  \textbf{GPT-2}
        &  \textbf{\makecell[l]{PG-TD}}
        &  \makecell[l]{27.873}
        &  \makecell[l]{23.226} 
        &  \makecell[l]{32.732} 
        &  \makecell[l]{2.200}
        &  \makecell[l]{8.367}
        &  \makecell[l]{\textbf{16.600}}
        \\
        &
        &  \textbf{\makecell[l]{ERP (Ours)}}
        &  \makecell[l]{\textbf{28.421}}
        &  \makecell[l]{\textbf{23.477}}
        &  \makecell[l]{\textbf{32.861}} 
        &  \makecell[l]{\textbf{2.300}}
        &  \makecell[l]{\textbf{8.567}}
        &  \makecell[l]{15.900}
        \\
        pass@15
        &  \textbf{GPT-Neo}
        &  \textbf{\makecell[l]{PG-TD}}
        &  \makecell[l]{28.928}
        &  \makecell[l]{24.267} 
        &  \makecell[l]{35.565} 
        &  \makecell[l]{2.400}
        &  \makecell[l]{\textbf{9.067}}
        &  \makecell[l]{19.000}
        \\
        &
        &  \textbf{\makecell[l]{ERP (Ours)}}
        &  \makecell[l]{\textbf{29.302}}
        &  \makecell[l]{\textbf{24.417}}
        &  \makecell[l]{\textbf{35.896}} 
        &  \makecell[l]{\textbf{2.500}}
        &  \makecell[l]{8.900}
        &  \makecell[l]{\textbf{19.700}}
        \\
        \midrule
        pass@20
        &  \textbf{GPT-2}
        &  \textbf{\makecell[l]{PG-TD}}
        &  \makecell[l]{28.142}
        &  \makecell[l]{23.977} 
        &  \makecell[l]{33.420} 
        &  \makecell[l]{2.300}
        &  \makecell[l]{8.700}
        &  \makecell[l]{16.800}
        \\
        &
        &  \textbf{\makecell[l]{ERP (Ours)}}
        &  \makecell[l]{\textbf{29.711}}
        &  \makecell[l]{\textbf{24.974}}
        &  \makecell[l]{\textbf{34.806}} 
        &  \makecell[l]{\textbf{2.400}}
        &  \makecell[l]{\textbf{9.200}}
        &  \makecell[l]{\textbf{17.000}}
        \\
        pass@20
        &  \textbf{GPT-Neo}
        &  \textbf{\makecell[l]{PG-TD}}
        &  \makecell[l]{29.209}
        &  \makecell[l]{24.992} 
        &  \makecell[l]{36.054} 
        &  \makecell[l]{2.400}
        &  \makecell[l]{9.233}
        &  \makecell[l]{19.300}
        \\
        &
        &  \textbf{\makecell[l]{ERP (Ours)}}
        &  \makecell[l]{\textbf{30.916}}
        &  \makecell[l]{\textbf{26.004}}
        &  \makecell[l]{\textbf{37.660}} 
        &  \makecell[l]{\textbf{2.700}}
        &  \makecell[l]{\textbf{9.433}}
        &  \makecell[l]{\textbf{20.400}}
        \\
        \bottomrule
    \end{tabular}}}
        \caption{\small \algname was compared to PG-TD for n@k and pass@k tasks (k=15, 20) on three APPS benchmarks (Introductory, Interview, Competition) using GPT-2 and GPT-Neo models. Evaluated by Pass Rate (PR) and Strict Accuracy (SA), \algname typically surpassed PG-TD.
        \label{exp:code}
        }

\end{table*}

\begin{table*}[ht]
\begin{minipage}[t]{0.5\linewidth}
\setlength{\tabcolsep}{4pt}
   \centering
        {\small
        \scalebox{0.75}{
        \begin{tabular}{llll}
        \toprule
              & 1@10            & 1@15            & 1@20            \\
        \midrule
        PG-TD & 21.438          & 23.242          & 23.540          \\
        ERP (Ours)  & \textbf{22.101}          & \textbf{23.955} & \textbf{25.204} \\
        \midrule
              & pass@10           & pass@15           & pass@20           \\
        \midrule
        PG-TD & 25.683          & 27.873          & 28.142          \\
        ERP (Ours)  & \textbf{25.958} & \textbf{28.421} & \textbf{29.711} \\
        \bottomrule
        \end{tabular}
        }
        \caption{\small The pass rate (\%) of 1@k and pass@k of ERP \((e=2)\)\\ and PG-TD, evaluated on the APPS introductory problems.  \label{tab:app-intro-comp}
        }
        
    }
\end{minipage}
\begin{minipage}[t]{0.5\linewidth}
\setlength{\tabcolsep}{4pt}
   \centering
        {\small
        \begin{tabular}{llll}
        \toprule
               & pass@10         & pass@15         & pass@20           \\
        \midrule
        PG-TD  & 25.683          & 27.873          & 28.142          \\
        ERP (\(e=2\)) & 25.958          & 28.421          & 29.711          \\
        ERP (\(e=3\)) & \textbf{26.073} & 28.712          & \textbf{29.991} \\
        ERP (\(e=4\)) & 25.840          & \textbf{28.745} & 29.990         \\
        \bottomrule
        \end{tabular}
        \caption{\small The pass rate (\%) of pass@k of ERP of varying forward steps \(e \in \{2,3,4\}\) and PG-TD on the APPS introductory problems.  \label{tab:erp_k}
        }
    }
\end{minipage}
\end{table*}

\subsubsection{Experimental configuration}

\paragraph{The language model.} Due to limited time and computational resources, we confined both the PG-TD and ERP methods to a rollout of 64, and further configured the ERP with an \(e\)-step of \(2\). We reused the GPT-x models from the PG-TD release versions. 

\paragraph{Dataset.} We conducted experiments on three benchmarks in total: APPS Intro, APPS Inter, APPS Comp from APPS \citep{hendrycks2021measuring}.%

\paragraph{Evaluation metric.} Performance was assessed using two metrics—pass rate and strict accuracy—and two tasks: n@k and pass@k \citep{li2022competition}. n@k measures the success rate of the top n selected programs in passing all private test cases. pass@k assesses the overall success rate of any of the k generated programs in passing all private test cases. For both ERP and PG-TD, the k samples consist of the full programs obtained from the initial k rollouts.
 
The pass rate represents the average percentage of private test cases that the generated programs successfully pass across all problems. Strict accuracy measures the percentage of problems for which the generated programs pass all private test cases. We report n@k and pass@k results with n and k values of 15 and 20, respectively. 

\subsubsection{Results}

\paragraph{On APPS task.} As shown in Table~\ref{exp:code}, our approach outperformed the current leading method, PG-TD, in all three code generation tasks across two LLM models, GPT-2 and GPT-Neo, on two metrics: pass rate and strict accuracy. We report 1@k and pass@k with k being 15 and 20. Our results demonstrate that the ERP method overall surpasses the PG-TD method.  In Table~\ref{tab:app-intro-comp}, we report more experimental results on APPS introductory problems with k from 10, 15, and 20, where ERP still performs better than the baseline.

\paragraph{Effect of \(e\)-step forward.}  The hyperparameter \(e\) in the forward step entropy reinforcement also has an effect on the performance on the task of code generation.  As can be seen from Table~\ref{tab:erp_k}, greater forward steps (3 or 4) lead to higher pass rates, which shows the importance of multi-step forward looking for our ERP algorithm.

\section{CONCLUSION}

\vspace{-0.2cm}
We introduce a novel algorithm, \algname, Entropy-Reinforced Planning for Transformer decoding. \algname employs an \emph{e}-step forward entropy-based MCTS planner to guide a Transformer decoder. Within the \algname framework, we incorporate the $\PHUCT$ algorithm for the selection phase and \text{TOP-PK} for the expansion phase of tree search planning. 
The resulting system: 
1) enhances sample efficiency by leveraging Transformer sampling to draw upon prior knowledge in the expansive molecule search space;
2) achieves controllable generation by adapting Transformer, which is trained on various objectives, and optimizes its decoding process to meet our specific goals; and 
3) reduces uncertainty and balances exploitation and exploration through entropy-reinforced planning, enhancing discovery of high-reward molecules in uncertain areas of molecular spaces. 
Empirical evaluations across a range of tasks demonstrate that \algname consistently surpasses the current state-of-the-art PG-TD and 
competing baselines. 
Our work highlights \algname's effectiveness in the domain of drug discovery \fix{ and code generation}, showing improvements in several pharmaceutical properties, \fix{ as well as pass rates and strict accuracy for code generation}. We encourage the application of \algname in domains beyond the scope of our current research.

\section*{Acknowledgements}
We thank Ian T. Foster for polishing the draft and Alex Brace for providing feedback on it. We also thank Martin L. Putra for the initial discussion.
This work is supported \fix{in part} by the RadBio-AI project (DE-AC02-06CH11357), U.S. Department of Energy Office of Science, Office of Biological and Environment Research, the Improve project under contract (75N91019F00134, 75N91019D00024, 89233218CNA000001, DE-AC02-06-CH11357, DE-AC52-07NA27344, DE-AC05-00OR22725), 
the Exascale Computing Project (17-SC-20-SC), a collaborative effort of the U.S. Department of Energy Office of Science and the National Nuclear Security Administration.

\section*{Impact Statement}
This paper presents work whose goal is to advance the field of machine learning for generation of structural sequences. 
A generative AI model for structural data has a few potential societal consequences, 
none of which we feel must be specifically highlighted here.

\bibliographystyle{plainnat}
\bibliography{reference}

\clearpage

\appendix

\onecolumn
\section{Appendix}

\subsection{Surrogate model}\label{app:surrogate_model}

We utilize a BERT-like transformer surrogate model for calculating docking scores, as described by \citet{vasan23} and \citet{liu23Drug}. 
Tokenized SMILES strings are input to the model and positionally embedded. The outputs are then processed through a series of five transformer blocks, each comprising a multi-head attention layer with 21 heads, a dropout layer, layer normalization with a residual connection, and a feed-forward network. 
The feed-forward network includes two dense layers, followed by dropout and layer normalization with a residual connection. 
Finally, the predicted docking score is output by the feed-forward network after the series of transformer blocks.

\subsection{Binding sites of RTCB and 3clpro}
\begin{figure*}[ht!]

    \begin{subfigure}{.5\textwidth}
        \centering
        \includegraphics[%
        width=4cm,  clip={0,0,0,0}]{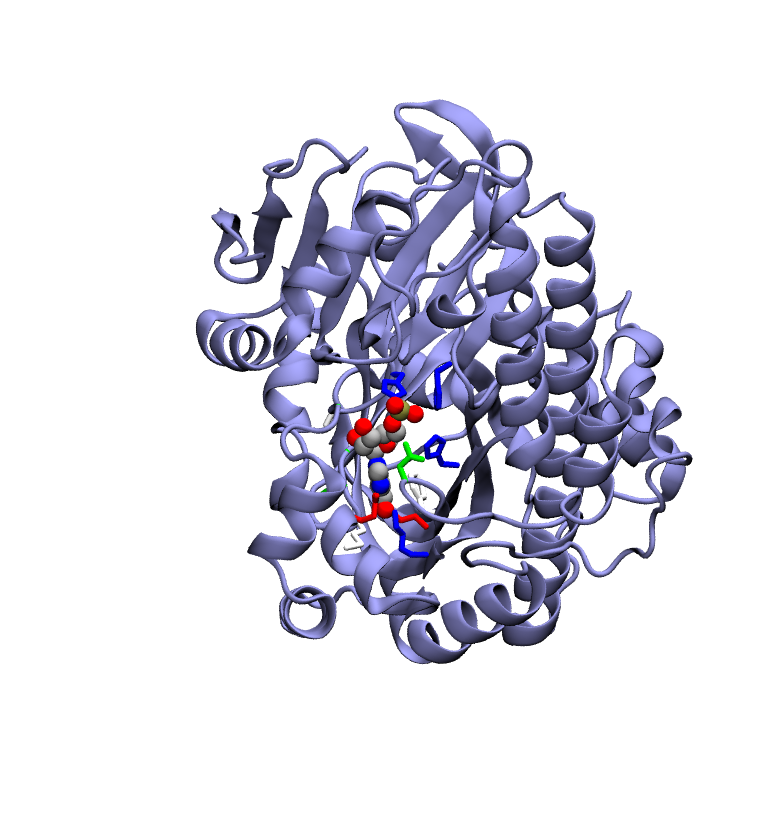}
        \caption{RTCB.}\label{fig:exp:sample}
    \end{subfigure}\hfil
    \begin{subfigure}{.5\textwidth}
        \centering
        \includegraphics[%
        width=4cm,  clip={0,0,0,0}]{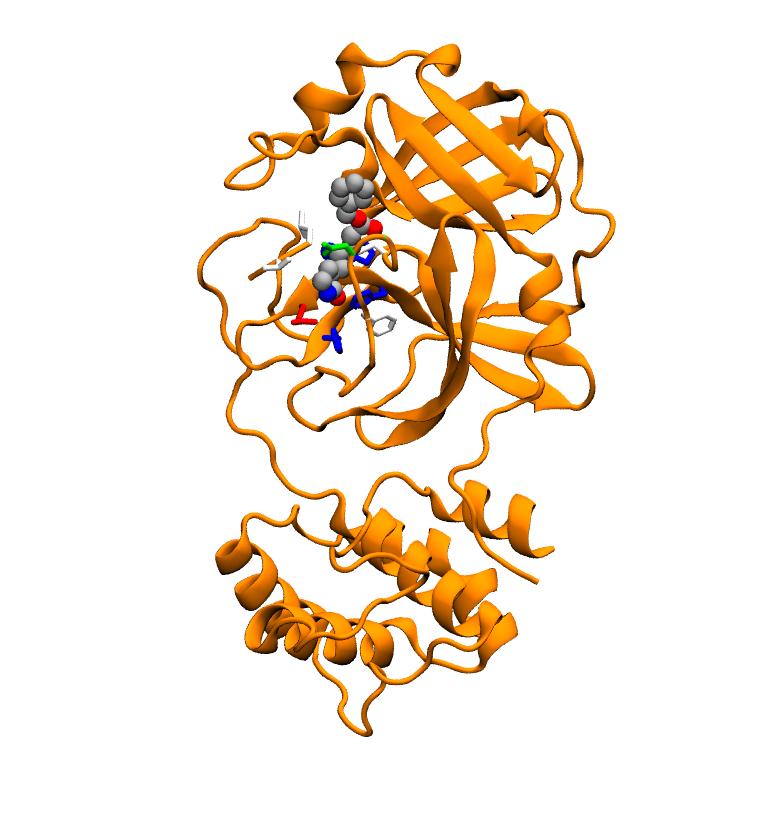}
        \caption{3CLPro.}\label{fig:exp:combine_expertise}
    \end{subfigure}\hfil

    \caption{The binding sites of RTCB (PDB ID: 4DWQ) and proteins 3CLPro (PDB ID: 7BQY). The Open Eye software is used to define binding sites surrounding the crystallized compound~\citep{kelley2015posit,liu23Drug}.     
    }
\end{figure*}

\fix{\subsection{Dataset detail}}\label{app:dataset}
\fix{Pretrained LLM is trained using about 10.7 million 
druglike and in-stock molecules from ZINC20 \citep{irwin2020zinc20} with standard reactivity. These molecules have a minimum sequence length of 8, an average length of 46, and a maximum length of 252.
The biased LLMs are finetuned from the pretrained LLM using the cancer and covid dataset of \citet{liu23Drug}. More specifically, 1 million compounds docked to 3CLPro~(PDB ID: 7BQY) protein associated with SARS-CoV-2 and the RTCB (PDB ID: 4DWQ) human cancer protein dataset with docking score in range [-14, -9], indicating string interactions, are selected for finetuning. 
When assessing the performance of our model, we go beyond solely checking if it can generate valid SMILES strings. Unlike code generation tasks, which are typically evaluated on a simple pass/fail basis depending on whether the code runs correctly, our evaluation for SMILES generation encompasses several additional pharmaceutical objectives: druglikeliness, docking score, synthesizability and solubility. This comprehensive approach ensures not only the generation of chemically valid structures but also evaluates their potential as effective drug candidates based on multiple critical factors.
}

\reb{\begin{table}[!h]
    \centering
    \begin{tabular}{|c|c|c|c|}
    \hline
        Dataset & Min Legnth & Mean Length & Max Length\\
        \hline
        ZINC & 8 & 46.0 & 252 \\
        \hline
        RTCB & 12 & 42.7 & 119 \\
        \hline
        3CLPro & 18&  49.6 & 107 \\
         \hline
    \end{tabular}
    \caption{Minimum, Mean and Maximum length of SMILES in each dataset}
    \label{tab:dataset_details}
\end{table}}
\XL{need update the properties of the dataset, desire attributes etc}

\subsection{Experiments on Code generation task}

In the main paper, we focus on addressing the drug discovery challenge, however the contribution of \algname is primarily methodological, which allows its application across various domains, including code generation. In this section, we conduct additional two experiments on two code generation dataset, APPS~\citep{hendrycks2021measuring} and CodeContests~\citep{li2022competition}. 
Here, we conducted an evaluation on base model GPT-2 and GPT-Neo using three additional code generation benchmark APPS dataset [1] including {APPS Intro, APPS Inter, APPS comp} and one CodeContests [2] datasets benchmark,  and compared our method to the existing state-of-the-art (PG-TD) on both pass rate and strict accuracy metric. Our approach surpassed the current leading method PG-TD on all the 4 code generation tasks across two LLMs model {GPT-2, GPT-Neo } on two metric {pass rate, Strict Accuracy} as well. Lastly, we have made updates to both papers (see appendix ??) and the codebase to reflect these new experiment results.

\subsection{Experiments on Molecules}

\subsubsection{Setup and hyperparameters}\label{app:setup}
For equitable assessment, we compare different algorithms based on similar setups.
In particular, in each comparison of different algorithms in all tables and charts, they share the key hyperparameters of the number of rollouts, and all generation and prediction configuration of the Transformer-based language models.

We did limited hyperparameter search based on the possible values in the range shown in Table~\ref{app:tab:hyperparams}.  Then we use the same hyperparameter to compare different algorithms.

\begin{table*}[ht!]
    {
    \centering
    {\footnotesize
{\begin{tabular}{@{}ll@{}}
\toprule
Hyperparameter  & Experimented values    \\ \midrule
{\makecell[l]{GPT2 Pretraining and Fine-tuning}} \\
{\makecell[l]{\quad Learning rate}} &  \makecell[c]{$5 \times 10^{-5}$} \\
{\makecell[l]{\quad Batch size per GPU}} &  \makecell[c]{8} \\
{\makecell[l]{\quad \# of GPU used}} &  \makecell[c]{8} \\
{\makecell[l]{\quad \# of epochs}} &  \makecell[c]{10} \\
{\makecell[l]{GPT2 Fine-tuning}} \\
{\makecell[l]{\quad \# of epochs}} &  \makecell[c]{40} \\
{\makecell[l]{Tree Search with ERP}} \\
\makecell[l]{\quad Number of rollouts \(N\)} & \makecell[c]{\(\{32, 64, 128, 256, 512\}\)}   \\ 
\makecell[l]{\quad Exploration parameter \(c_p\)}             &          \makecell[c]{\(\{1, 4, 8\}\)}        \\
\makecell[l]{\quad LM top-p filter for expansion \(p\)} & \makecell[c]{\(\{0.9, 0.95\}\)} \\
\makecell[l]{\quad LM top-k filter for expansion \(k\)} & \makecell[c]{\(\{15, 20\}\)} \\
\makecell[l]{\quad LM beam size for evaluation \(b\)} & \makecell[c]{\(\{8, 16\}\)} \\ 
\makecell[l]{\quad Forward step \(e\)} & \makecell[c]{\(\{1, 2, 3, 4, 5, 6\}\)} \\ \bottomrule
\end{tabular}}

        \caption{{{Hyperparameters}} of possible values.  We made limited search among the possible values and compare different algorithms based on the same applicable hyperparameters.}
        \label{app:tab:hyperparams}
        }}
\end{table*}

\subsubsection{Computing infrastructure {and wall-time comparison}}\label{app:computing_infrastructure}
We trained our docking surrogate models using four nodes of the \fix{supercomputer} where each node contains one 64-core CPU and four A100 GPU nodes~\citep{Polaris}.
The training time for each model was approximately three hours.
We conducted the Monte-Carlo tree search with Transformer inference on a cluster that includes CPU nodes and GPU nodes two Nvidia GPUs. {Based on the computing infrastructure, we obtained the wall-time comparison in \tabref{table:wall-time} as follows.}

 \begin{table*}[ht!]
             \centering
 {
    {\footnotesize
    \scalebox{1}{
    \begin{tabular}{l c c  }
        \toprule
        \textbf{Methods}
        & {\makecell[c]{Total Run Time}}
        \\
        \midrule
        \textbf{\makecell[l]{Pretrain GPT}}
        &  \makecell[r]{9 hours}
        \\
        \textbf{\makecell[l]{Biased GPT}}
        &  \makecell[r]{17 hours}
        \\
        \textbf{\makecell[l]{RL finetuned GPT}}
        &  \makecell[r]{17 hours}
        \\
        \midrule
        \textbf{\makecell[l]{Sampling}}
        &  \makecell[r]{10 min.}
        \\
        \textbf{\makecell[l]{PG-TD}}
        &  \makecell[r]{30 min.}
        \\
        \textbf{\makecell[l]{ERP}}
        &  \makecell[r]{40 min.}
        \\
        \bottomrule
    \end{tabular}}}
        \caption{{Wall-time comparison between different methods.} }
        \label{table:wall-time}  
        }
\end{table*}

\subsubsection{Extended baselines}\label{app:sec:extendedbaselines}
We present results with some more baselines in Table~\ref{app:tbl:extendedbaselines}, which demonstrated the complexity of the task of molecule generation.  Specifically, it provides the UCT-only baseline, which cannot generate any valid molecules in our experimentation.  
Furthermore, using LM in the expansion phase but not in the selection has minuscule improvement over random sampling baseline, and shows the importance of using the LM in the selection phase as well.
\begin{table}[ht]
\footnotesize
\setlength{\tabcolsep}{8pt}
\centering
\begin{tabular}{@{}llll@{}}
\toprule
Model                                               & Avg. norm. reward & Ratio of valid mol. & \# unique valid mol. found \\ \midrule
Uniform random sampling                         & Null                      & 0.0                      & 0                                      \\
UCT without LM                                  & Null                      & 0.0                      & 0                                      \\
UCT using LM in expansion only                  & 1.99                      & 0.0001                   & 1                                      \\
Beam search with pre-trained LM (as in Table~\ref{exp:main_result}) & 2.38                      & 1.0                      & 16                                     \\
Sampling with pre-trained LM (as in Table~\ref{exp:main_result})    & 2.39                      & 1.0                      & 1609                                   \\
UCT (as in Table~\ref{exp:main_result})                             & 2.25                      & 1.0                      & 3013                                   \\
PG-TD (as in Table~\ref{exp:main_result})                           & 2.59                      & 1.0                      & 2549                                   \\
ERP (as in Table~\ref{exp:main_result})                             & 2.62                      & 1.0                      & 2575                                   \\ \bottomrule
\end{tabular}
\caption{\label{app:tbl:extendedbaselines}
Results with some more baselines with the same setting as in Table~\ref{exp:main_result}, which demonstrated the complexity of the task of molecule generation.}
\end{table}

\subsubsection{More ablations}\label{app:sec:moreablations}
\paragraph{Token-budgeted comparisons}  In Table~\ref{app:tab:token-budgeted}, we show the performance of ERP and PG-TD with upper-bounded budget of the number of generated tokens by the LM in the tasks of molecular generation.  It shows that ERP has a stronger performance than PG-TD given a LM-generated token budget, given number of rollouts being 32.
\begin{table}[ht]
\footnotesize
\setlength{\tabcolsep}{8pt}
\centering
\begin{tabular}{@{}llll@{}}
\toprule

Number of sampled tokens & \multicolumn{2}{c}{Average normalized reward} &  \\
\cmidrule(lr){2-3}
   & ERP with e-etep=6 & PG-TD &  \\ \midrule
1024                     & 2.36              & 2.36  &  \\
2048                     & 2.28              & 2.25  &  \\
4096                     & 2.34              & 2.28  &  \\
8192                     & 2.39              & 2.35  &  \\
16384                    & 2.54              & 2.43  &  \\
32768                    & 2.60              & 2.57  &  \\
65536                    & 2.60              & 2.57  &  \\ \bottomrule
\end{tabular}
\caption{\label{app:tab:token-budgeted} Average normalized rewards given number of rollouts being 32, and varying budget of sampled tokens.}
\end{table}

\paragraph{Effects of Top-pk expansion}
In Table~\ref{app:tab:expansion} we show the ablation of removing Top-pk filtering during the expansion phase. The ablation results between Row~1 and Row~4 show the importance of top-pk filtering in ERP, whereas the ablation results between Row~2 and Row~4 show the effects of ERP.
\begin{table}[ht]
\footnotesize
\setlength{\tabcolsep}{8pt}
\centering
\begin{tabular}{@{}llll@{}}
\toprule
E-step (Selection) & Top-P-K filtering (Expansion) & Average reward & Note                                   \\ \midrule
6                  & False                         & 2.46           &                                  \\
0                  & True                          & 2.57           & Reduced to P-UCB, As in Figure 3 (h) \\
1                  & True                          & 2.57           & As in Figure 3 (h)                   \\
6                  & True                          & 2.60           & As in Figure 3 (h)                   \\ \bottomrule
\end{tabular}
\caption{\label{app:tab:expansion} Average normalized rewards showing the effects of top-pk filtering during expansion.}
\end{table}

\end{document}